%% file: main.tex
\documentclass{article}
\PassOptionsToPackage{hyphens}{url}
\usepackage[preprint]{icml2026}
\usepackage[utf8]{inputenc} 
\usepackage[T1]{fontenc}    
\usepackage{hyperref}       
\usepackage{url}            
\usepackage{booktabs}       
\usepackage{amsfonts}       
\usepackage{nicefrac}       
\usepackage{microtype}      
\usepackage{xcolor}         
\usepackage{graphicx}
\usepackage{caption}
\usepackage{subcaption}
\usepackage{enumitem}
\setlist[enumerate]{nosep}

\usepackage{amsmath, amsthm, amssymb}
\usepackage[capitalize,noabbrev]{cleveref}
\usepackage[textsize=tiny]{todonotes}
\usepackage{rotating}

\makeatletter
\AddToHook{cmd/appendix/before}{%
    \crefalias{section}{appendix}%
    \crefalias{subsection}{appendix}
}
\makeatother

\newtheorem{theorem}{Theorem}[section]
\theoremstyle{definition}
\newtheorem{definition}[theorem]{Definition}
\theoremstyle{remark}

\newcommand{\R}{\mathbb{R}}
\DeclareMathOperator{\Agg}{Agg}
\DeclareMathOperator{\Up}{Up}
\DeclareMathOperator{\MSE}{MSE}

\DeclareMathOperator{\Unif}{Unif}
\DeclareMathOperator{\ReLU}{ReLU}

\newcommand{\transpose}{\intercal}
\newcommand{\abs}[1]{\left| #1 \right|}
\newcommand{\partials}[2][]{\frac{\partial #1}{\partial #2}}

\icmltitlerunning{Mechanistic Interpretability for Neural Algorithmic Reasoning}

\begin{document}

\twocolumn[
  \icmltitle{MINAR: Mechanistic Interpretability for Neural Algorithmic Reasoning}

  \icmlsetsymbol{equal}{*}

  \begin{icmlauthorlist}
    \icmlauthor{Jesse He}{pnnl,ucsd}
    \icmlauthor{Helen Jenne}{pnnl}
    \icmlauthor{Max Vargas}{pnnl}
    \icmlauthor{Davis Brown}{pnnl,penn}
    \icmlauthor{Gal Mishne}{ucsd}
    \icmlauthor{Yusu Wang}{ucsd}
    \icmlauthor{Henry Kvinge}{pnnl,uw}
  \end{icmlauthorlist}

  \icmlaffiliation{pnnl}{Pacific Northwest National Laboratory, Richland, WA}
  \icmlaffiliation{ucsd}{Halıcıoğlu Data Science Institute, University of California, San Diego, San Diego, CA}
  \icmlaffiliation{penn}{Department of Computer and Information Science, University of Pennsylvania, Pennsylvaina, PA}
  \icmlaffiliation{uw}{Department of Mathematics, University of Washington, Seattle, WA}

  \icmlcorrespondingauthor{Jesse He}{jeh020@ucsd.edu}
  \icmlkeywords{}

  \vskip 0.3in
]

\printAffiliationsAndNotice{}

\input{sec/0-abstract}
\input{sec/1-introduction}
\input{sec/2-related_work}
\input{sec/3-preliminaries}
\input{sec/4-method}
\input{sec/5-bellman_ford}
\input{sec/6-salsa-clrs}
\input{sec/7-conclusion}
\input{sec/ack}
\input{sec/impact}

\bibliographystyle{plainnat}
\bibliography{references}
\newpage

\onecolumn
\appendix
\input{sec/appdx/additional-bf-results}
\newpage
\input{sec/appdx/additional-clrs-results}
\newpage
\input{sec/appdx/dataset}
\input{sec/appdx/training}

\end{document}

%% file: sec/0-abstract.tex
\begin{abstract}
    The recent field of \emph{neural algorithmic reasoning} (NAR) studies the ability of graph neural networks (GNNs) to emulate classical algorithms like Bellman-Ford, a phenomenon known as \emph{algorithmic alignment}.
    At the same time, recent advances in large language models (LLMs) have spawned the study of \emph{mechanistic interpretability}, which aims to identify granular model components like \emph{circuits} that perform specific computations.
    In this work, we introduce \textbf{\underline{M}}echanistic \textbf{\underline{I}}nterpretability for \textbf{\underline{N}}eural \textbf{\underline{A}}lgorithmic \textbf{\underline{R}}easoning (MINAR), an efficient circuit discovery toolbox that adapts attribution patching methods from mechanistic interpretability to the GNN setting. We show through two case studies that MINAR recovers faithful neuron-level circuits from GNNs trained on algorithmic tasks. Our study sheds new light on the process of circuit formation and pruning during training, as well as giving new insight into how GNNs trained to perform multiple tasks in parallel reuse circuit components for related tasks.
    Our code is available at \url{https://github.com/pnnl/MINAR}.
\end{abstract}

%% file: sec/1-introduction.tex
\section{Introduction}
\label{sec:introduction}

The recent surge in the capabilities of large language models (LLMs) has created a commensurate demand for novel interpretability methods suited to rigorous analysis of these models. At the same time, the discovery of phenomena like \emph{grokking}~\cite{power2022grokkinggeneralizationoverfittingsmall} has shown that rich algorithmic structure can arise via standard training methods. This confluence of interests has led to the emergence of \emph{mechanistic interpretability}~\cite{nanda2023progress}, which attempts to explain model behavior using specific subsets of a model's internal computation. In several striking cases \cite{wang2023interpretability, lindsey2025biology} this has allowed researchers to reduce seemingly complex behaviors to an interpretable sequence of mathematical operations.

At the same time, researchers have noticed that graph neural networks (GNNs) are capable of implementing classical algorithms as a result of the similarity between their message-passing structure and dynamic programming. This notion of ``algorithmic alignment''~\cite{Xu2020What} or ``neural algorithmic reasoning'' ~\cite{velickovic2020Neural}, promises robust out-of-distribution generalization and substantial improvements in sample complexity. With an interest in discovering how models perform algorithmic tasks on the one hand, and a class of models (GNNs) that are known to naturally implement traditional algorithms on the other, neural algorithmic reasoning creates a natural testbed for fine-grained mechanistic interpretability research.

In this work, we introduce \textbf{\underline{M}}echanistic \textbf{\underline{I}}nterpretability for \textbf{\underline{N}}eural \textbf{\underline{A}}lgorithmic \textbf{\underline{R}}easoning (MINAR). To our knowledge, MINAR is the first attempt to apply automated circuit discovery to the GNN setting.
Through two case studies, we demonstrate how MINAR enables a detailed study of neural algorithmic reasoning models.
Our contributions are as follows:
\begin{enumerate}
    \item Building on prior work in mechanistic interpretability, we develop an efficient approach for identifying end-to-end neuron-level circuits in GNNs (\cref{sec:method}).
    \item Applied to a GNN trained to predict single-source shortest path distances, we show that MINAR extracts a minimal circuit that captures the model's implementation of the Bellman-Ford algorithm (\cref{subsec:bellman-ford-circuit}). 
    \begin{enumerate}
        \item MINAR also reveals a delayed pruning effect, where the optimal circuit appears to be pruned from a larger circuit long after a generalizable model has been obtained.
        (\cref{subsec:circuit-grokking}).
        \item MINAR exposes shortcut learning in a network trained to perform Bellman-Ford and breadth-first search (BFS) in parallel (\cref{subsec:bf-bfs-parallel}). 
    \end{enumerate}
    \item We study a GNN trained to perform seven tasks from the CLRS-30 benchmark \cite{velickovic2022clrs} in parallel. Here, MINAR showcases how the model leverages the same circuit components to perform related tasks (\cref{sec:salsa-clrs}).
\end{enumerate}

%% file: sec/2-related_work.tex
\section{Related Work}
\label{sec:related-work}

\noindent \textbf{Circuit Discovery and Mechanistic Interpretability}
Early mechanistic interpretability work, e.g.,~\cite{nanda2023progress, zhong2023clock}, combines manual examination of parameters with expert hypotheses to reverse-engineer the inner workings of a narrow trained model. However, the rise in language model capabilities has created an interest in applying such methods to much larger models, focusing on methods to partially or fully automate much of the mechanistic interpretability process. In particular, researchers have identified ``circuits'' that perform specific subtasks \cite{olah2020zoom}, and the process of identifying circuits is known as \emph{circuit discovery}. A number of approaches for circuit discovery have been proposed, such as attribution patching~\cite{syed2023attributionpatchingoutperformsautomated, hanna2024have, zhang2025eapgp} or pruning~\cite{cao-etal-2021-low, conmy2023acdc, yu2024functionalfaithfulnesswildcircuit}. These methods formulate the model as a \emph{computation graph} and attempt to find small subgraphs responsible for a particular behavior. To date, much of the circuit discovery literature has been focused on sub-tasks for language models such as indirect object identification \citep{wang2023interpretability}, and has identified coarse-grained circuits whose nodes are entire attention heads or MLP modules.

\noindent \textbf{Algorithmic Alignment in Graph Neural Networks}
While mechanistic interpretability has revealed surprising structure in large generalist models, a different class of narrow models has also been shown to possess algorithmic ``reasoning'' capabilities: graph neural networks (GNNs). Broadly, GNNs operate via an iterative \emph{message-passing} scheme: each node in the graph maintains an embedding, and iteratively aggregates the embeddings of its neighbors to update its own embedding. \citet{Xu2020What} and \citet{dudzik2022graph} point out that this aggregate-and-update flow resembles dynamic programming, hypothesizing that this \emph{algorithmic alignment} of GNNs affords them an advantage in learning algorithmic tasks where more ad hoc approaches might fail. For example, learning reachability via BFS and shortest paths via Bellman-Ford in parallel can help GNNs generalize on both tasks~\cite{velickovic2020Neural}. \citet{velickovic2022clrs} later introduced the CLRS-30 benchmark to demonstrate this parallel reasoning ability across 30 algorithmic tasks inspired by the classic textbook \textit{Introduction to Algorithms} by Cormen, Leiserson, Rivest, and Stein \cite{cormen2022introduction}. However, it was only recently that \citet{nerem2025graphneuralnetworksextrapolate} showed that GNNs are not only \emph{capable} of learning specific algorithms, but that a properly designed GNN will \emph{provably} learn a specific algorithm (in their case Bellman-Ford) when trained with sparsity regularization.

%% file: sec/3-preliminaries.tex
\section{Preliminaries}
\label{sec:preliminaries}
Let $\Phi$ be a GNN operating on an attributed graph $G = (V, E, X, A)$, where $V$ is the node set, $E$ is the edge set, $X : V \to \R^p$ are the node features and $A : E \to \R^q$ are edge attributes.
We briefly recall the message-passing scheme that underlies most popular GNNs. A GNN $\Phi$ of depth $L$ maintains an embedding $\Phi^{(\ell)}_v$ for each node $v \in V$ and $\ell = 0, \dots, L$, with layer 0 being the input node features $\Phi^{(0)}_v = X_v$. In each layer, $\Phi^{(\ell)}_v$ is updated according to the features of its 1-hop neighborhood $\mathcal{N}(v)$:
\begin{equation}
    \Phi^{(\ell+1)}_v = f^{(\ell)}_{\Up}\left(\Phi^{(\ell)}_v,
        \bigoplus_{u \in \mathcal{N}(v)} f^{(\ell)}_{\Agg}(\Phi^{(\ell)}_u, A_{(u,v)})
    \right)
    \label{eq:mpnn}
\end{equation}
where $f^{(\ell)}_{\Agg}$ and $f^{(\ell)}_{\Up}$ are any functions and $\bigoplus$ denotes any permutation-invariant operation such as sum, mean, minimum, or maximum. We may denote the output of $\Phi$ at node $v$ by $\Phi_v$ or $\Phi_v(G)$, if the input graph is ambiguous.

Fundamental to circuit discovery is the \emph{computation graph}, which represents the connections between individual model components~\cite{olah2020zoom, syed2023attributionpatchingoutperformsautomated, zhang2025eapgp}\footnote{
    In GNN literature, some authors~\cite{ying2019gnnexplainer, you2021identity, he2025explaining} use ``computation graph'' to refer to a GNN's layered message-passing, but in this paper ``computation graph'' will always refer to the model computation graph.
}. These components are typically defined at an architectural level, representing modules like attention heads or MLPs.
Here, we will denote by $\Psi$ an arbitrary (perhaps non-graph) neural network.
\begin{definition}[Model Computation Graph]
    Let $\Psi$ be a neural network with arbitrary subcomponents $\psi_i$ (e.g. neurons). The \emph{model computation graph} of $\Psi$, denoted $G_c(\Psi)$, is the directed graph with vertices $\psi_i$ and edges $(i, j)$ if the output of $\psi_i$ is part of the input to $\psi_j$.
\end{definition}
While prior work in LLMs frequently defines these components at a lower resolution (e.g., entire attention heads or MLPs), our work is more in line with the neuron-level circuits that \citet{olah2020zoom} identify in convolutional neural networks: each node in the computation graph corresponds to an individual neuron in the model, and each edge corresponds to a connection between neuron activations.

We also clarify here a subtle point for the model computation graph of a GNN: because a message-passing GNN \eqref{eq:mpnn} shares parameters across each node in the input graph, any neuron $\psi$ will necessarily have several activations on a single graph instance. Therefore, fully unrolling the computation graph would require a copy of $\psi$ for every node in the input graph. This differs from circuit discovery in the LLM setting, e.g.~\cite{wang2023interpretability}, where the circuit can receive input directly from the model's residual stream and the circuit output is a single next-token prediction. Since our goal will be to extract a circuit which is shared across graphs, we will consider only a single copy of each neuron. We discuss in the next section how we accomodate the multiplicity of each neuron in the GNN computation graph.

%% file: sec/4-method.tex
\section{Circuit Discovery for GNNs}
\label{sec:method}

In \cref{subsec:attribution-patching} we first describe the scoring methods edge attribution patching (EAP)~\cite{syed2023attributionpatchingoutperformsautomated} and edge attribution patching with integrated gradients (EAP-IG)~\cite{hanna2024have}.\footnote{
We also note the more recent method EAP-GP~\cite{zhang2025eapgp}, although we will not discuss it in detail here.}
We then describe our main technical contribution: adapting mechanistic interpretability methods to extract circuits in the GNN setting (\cref{subsec:gnn-circuit-discovery}).

\subsection{Attribution Patching Scores}
\label{subsec:attribution-patching}

Early circuit discovery work employs \emph{activation patching}, which removes an edge from the computation graph and computes how this perturbation affects the model's predictions.~\cite{meng2022locating, conmy2023acdc}.
Formally, given a loss $\mathcal{L}$ and input $x$, let $(i,j)$ be an edge in the computation graph of a neural network $\Psi$, corresponding to the connection between two modules $\psi_i$ and $\psi_j$. Let $\Psi^{\setminus (i,j)}$ be the neural network $\Psi$ with computation edge $(i,j)$ removed. Then $(i,j)$ can be given an activation patching score
\begin{equation}
    \textsc{ActPatch}(i,j) = \mathcal{L}\left(\Psi(x), \Psi^{\setminus (i,j)}(x)\right).
\end{equation}
In the LLM setting, $\mathcal{L}$ may be the absolute logit difference for the ``clean'' output token~\cite{syed2023attributionpatchingoutperformsautomated} or the KL divergence between two token distributions~\cite{conmy2023acdc}.
Activation patching can be seen as a causal intervention on the model behavior. However, it becomes prohibitively expensive for larger computation graphs, as it requires a forward pass for each computation edge.

The goal of \emph{attribution patching} is to \emph{approximate} this difference in prediction~\cite{hanna2024have}.
Now, for a computation edge $(i,j)$, we assign an attribution score of $(i,j)$ for a prediction $\Psi(x)$ as follows: we first create a corrupted input $x'$. In the LLM setting, $x'$ is typically a sentence where tokens of interest like names or years have been changed, but the overall sentence structure is preserved.
Then, given the activation $z_i$ at $\psi_i$ for the input $x$, the activation $z_i'$ from the corrupted input $x'$, and the loss $\mathcal{L}$ between predictions $y = \Psi(x)$ and $y' = \Psi(x')$, the EAP score uses the gradient of $\mathcal{L}$ with respect to the output of $\psi_j$. That is,
\begin{equation}
    \operatorname{EAP}_{(i,j)}(x,x') = (z_i'-z_i)^\transpose \partials{\psi_j}
    \mathcal{L}(\Psi(x),\Psi(x')).
    \label{eq:EAP}
\end{equation}
Equation \eqref{eq:EAP} comes from the first-order term in a Taylor expansion for the perturbation performed by activation patching~\cite{syed2023attributionpatchingoutperformsautomated}, similar to the relationship between occlusion search and {Input $\odot$ Gradient} described by~\cite{ancona2018towards} in classical explainability. By perturbing the input and using the intermediate activations, EAP scores can be computed more efficiently than activation patching scores: EAP performs just two forward passes and one backward pass for each pair of inputs, rather than a forward pass for each computation edge~\cite{syed2023attributionpatchingoutperformsautomated}.

Extending the analogy between circuit discovery and classical explainability, EAP-IG~\cite{hanna2024have} adapts EAP to use the integrated gradients method of~\citet{sundarajan2017axiomatic}, inheriting its advantages over Input$\odot$Gradient. EAP-IG approximates an integral over the straight-line path between $z_i$ and $z_i'$ with a Riemann sum of $m$ terms:
\begin{equation}
\begin{split}
    &\operatorname{EAP-IG}_{(i,j)}(x,x') = \\
    &(z_i'-z_i)^\transpose \frac{1}{m}\sum_{k=1}^m \partials{\psi_j}
    \mathcal{L}\left(\Psi(x),\Psi\left(x'+\frac{k}{m}(x-x')\right)\right).
\end{split}
    \label{eq:EAP-IG}
\end{equation}
EAP-IG is $m$ times slower than EAP, essentially performing the EAP calculation $m$ times~\cite{hanna2024have}.

\subsection{Circuit Identification in GNNs}
\label{subsec:gnn-circuit-discovery}

Having introduced EAP and EAP-IG, we now turn to one of the central contributions of this work: our adaptation of attribution patching to the graph neural network setting.
Because a graph neural network operates on each node of the input graph, each computation edge may receive several scores for a single prediction, corresponding to each node of the underlying input graph. To give each computation edge a single score for a given prediction, in MINAR we take the mean over the scores given by each of these nodes.

That is, given a (clean) input graph $G$, we design a corrupted input graph $G'=(V,E,X',A')$. Note that $G'$ will have the same node set $V$ and edge set $E$ as $G$, but will have corrupted features $X'$ and $A'$. This way, $G'$ remains structurally aligned with $G$, allowing for a one-to-one comparison of node and edge activations while ablating the actual computation. (We discuss this choice further in \cref{appdx:corr-data}.) Then given a GNN $\Phi$, we define
\begin{equation}
    \begin{split}
        &\operatorname{EAP}_{(i,j)}(G,G') = \\
        &\phantom{-}\frac{1}{|V|}\sum_{v \in V} \left({{z'}^v_i}-z_i^v\right)^\transpose \partials{\psi_j}
        \mathcal{L}(\Phi_v(G),\Phi_{v}(G'))
    \end{split}
    \label{eq:graph-eap}
\end{equation}
where $z_i^{v}$ is the activation of neuron $\psi_i$ on node $v$ and ${z_i'}^{v}$ is the corrupted activation of $\psi_i$ on $v$. We note that the mean in \eqref{eq:graph-eap} can be replaced by any pooling operation.
We compute EAP-IG in a similar manner, again taking the mean score over each node. For EAP-IG, we implement the steps between the original and corrupted inputs by interpolating between all node and edge features.

Finally, since attribution scores are typically averaged over multiple input instances in order to understand their behavior across a probing dataset $\mathcal{G}_{\text{probe}}$, we average over each input graph to obtain the final attribution score for each edge. We also follow previous work in using the absolute value of each attribution score for downstream circuit discovery:
\begin{equation}
    \operatorname{EAP}_{(i,j)} = \abs{
    \frac{1}{|\mathcal{G}_{\text{probe}}|} \sum_{(G,G') \in \mathcal{G}_{\text{probe}}}
    \operatorname{EAP}_{(i,j)}(G,G')
    }
\end{equation}
and similarly for EAP-IG. We note that the mean in \eqref{eq:graph-eap} can be replaced by any pooling operation (e.g. sum).

Here, we also introduce two simple but natural baseline scoring methods. First, since computation edges represent connections between individual neurons, we can simply use the weight $W_{i,j}$ between neurons $\varphi_i$ and $\varphi_j$ as a score function. This simple baseline is invariant to specific tasks or perturbations, and is therefore unsuitable for isolating specific behaviors in models performing multiple algorithmic computations in parallel.
For the second baseline, our neuron-level approach also admits a simple sensitivity analysis by taking the gradient of the loss between clean and corrupted outputs with respect to the weight:
\begin{equation}
    \textsc{WeightGrad}(i,j) = \frac{\partial}{\partial W_{i,j}} \mathcal{L}(\Phi(x),\Phi(x')).
\end{equation}
These scores are normally ill-defined in the LLM setting, where computation edges represent large collections of weights, but they are natural choices at the neuron level.

To guarantee \emph{connectedness} of the identified circuit, we construct the circuit from complete paths between the computation graph's inputs and outputs, roughly following the ``path patching'' approach of~\cite{wang2023interpretability}.
Because we wish to maximize the attribution scores across the circuit, MINAR takes advantage of the fact that the computation graph is directed and acyclic, and therefore supports an efficient longest-path algorithm. Each longest path computation takes $O(|V(G_c)| + |E(G_c)|)$ time, the same runtime as constructing the model computation graph $G_c$ itself.

\begin{algorithm}
  \caption{Circuit construction algorithm}
  \label{alg:circuit_discovery}
  \begin{algorithmic}
    \STATE {\bfseries Input:} $G_c$, \texttt{Scores}, $K$ 
    \STATE \textsc{TopologicalSort}($G_c$)
    \STATE \textsc{SortDescending}(\texttt{Scores})
    \STATE Initialize circuit $C$
    \STATE $k \gets 0$
    \WHILE{$k < K$ and $\texttt{Scores} \neq \varnothing$}
        \STATE $(i,j) \gets \texttt{Scores.}\textsc{Pop}()$
        \IF{$(i,j) \notin C$}
            \STATE $C \gets C \ \cup$ \textsc{LongestPathWithEdge}($G_c, (i,j)$)
            \STATE $k \gets k + 1$
        \ENDIF
    \ENDWHILE
    \RETURN $C$
  \end{algorithmic}
\end{algorithm}
\cref{alg:circuit_discovery} describes our circuit discovery algorithm. Once scores are computed and the computation graph is constructed, topologically sorting $G_c$ takes time $O(|V(G_c)| + |E(G_c)|)$, and sorting the edges by attribution score takes time $O(|E(G_c)|\log(|E(G_c)|))$. After these preprocessing steps, finding a circuit that includes the top $K$ edges takes time $O(K(|V(G_c)| + |E(G_c)|))$. 

By constructing the circuit from complete input-output paths, MINAR guarantees that the identified circuit is a connected subgraph of the computation graph whose only parentless and childless nodes are the model inputs and outputs, respectively. This differs from previous circuit discovery work, which uses naive top-$k$ selection~\cite{syed2023attributionpatchingoutperformsautomated} or a greedy Dijkstra-like algorithm~\cite{hanna2024have}. These methods often produce disconnected circuits or circuits with parentless or childless edges which must then be pruned. In contrast, MINAR will always produce a connected end-to-end circuit, ensuring that the result can be applied as a subnetwork of the original network.

%% file: sec/5-bellman_ford.tex
\section{Case Study: Bellman-Ford}
\label{sec:bellman-ford}

We validate MINAR by studying GNNs trained to perform single-source shortest path distances, a common task for neural algorithmic reasoning research~\cite{velickovic2020Neural, Xu2020What, nerem2025graphneuralnetworksextrapolate}. MINAR recovers the minimal circuit in a GNN trained to compute single-source shortest path differences (\cref{subsec:bellman-ford-circuit}). We also present a surprising observation that the parameters of the Bellman-Ford circuit generalize later than the full model (\cref{subsec:circuit-grokking}). \cref{subsec:bf-bfs-parallel} extends this example by training on a second task, breadth-first search (BFS), in parallel. Here, applying MINAR reveals that model learned a shortcut for breadth-first-search based on the shortest path distance. Although we report here results from a single run, we observe the same behavior from different initializations (\cref{appdx:seeds}) and experimental setups (\cref{appdx:no-self-loops}).

\subsection{Training}

Throughout, we replicate the training setup of~\cite{nerem2025graphneuralnetworksextrapolate}.
We use a two-layer minimum-aggregated message-passing network (MinAggGNN), corresponding to two steps of Bellman-Ford. Recall that at the $\ell$-th step of the Bellman-Ford algorithm, each node $v$ updates its estimated distance $d_v^{(\ell)}$ to the source node $s$ by
\begin{equation}
    d_v^{(\ell+1)} \gets \min_{u \in \mathcal{N}(v)}\{ d_u^{(\ell)} + w_{u,v} \}
    \label{eq:bellman-ford}
\end{equation}
where $w_{u,v}$ is the weight of edge $(u,v)$. Analogously, each layer of the MinAggGNN is given by
\begin{equation}
    \Phi^{(\ell+1)}_v = f^{(\ell)}_{\Up}\left(\Phi^{(\ell)}_v,
        \min_{u \in \mathcal{N}(v)} f^{(\ell)}_{\Agg}(\Phi^{(\ell)}_u, e_{u,v})
    \right)
    \label{eq:minagg-gnn}
\end{equation}
where $e_{u,v}$ is the edge feature for $(u,v)$. We implement $f^{(\ell)}_{\Agg}$ and $f^{(\ell)}_{\Up}$ as two-layer MLPs. By using minimum aggregation, we see that the GNN only needs to learn a simple inner function to emulate \cref{eq:bellman-ford}. 
By training with $L_1$ sparsity regularization and a small curated training set,~\cite{nerem2025graphneuralnetworksextrapolate} showed that such a GNN must implement the Bellman-Ford algorithm.

This training set consists of few small graphs of size up to 6 nodes. To ensure that the test set captures out-of-distribution generalization, the test set consists of hundreds of randomly generated graphs up to 200 nodes. Nodes are given input features that emulate the standard initialization of the Bellman-Ford algorithm from some source node $s$:
\begin{equation}
    x_v^{\text{SP}} = \begin{cases}
        0 & v \text{ is the source} \\
        B & \text{otherwise}.
    \end{cases}
    \label{eq:sp-encoding}
\end{equation}
Here $B$ is any large number greater than any path length, which we take to be $B = 1000$. We discuss the dataset in greater detail in \Cref{appdx:sp-data}.

\begin{figure}
    \centering
    \includegraphics[width=\linewidth]{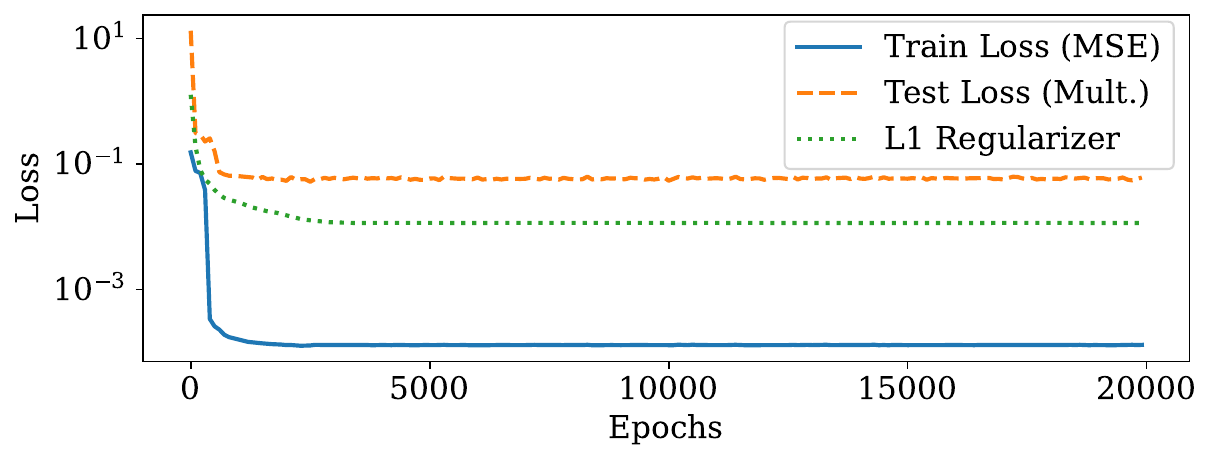}
    \caption{MSE training loss, multiplicative test loss, and $L_1$ regularization term for Bellman-Ford MinAggGNN.}
    \label{fig:shortest_path_loss}
\end{figure}

While we use MSE during training, we evaluate on the test set $\mathcal{G}_{\text{test}}$ using a multiplicative test loss $\mathcal{L}_{\text{Mult}}$ which is not sensitive to the scale of the edge weights. We compute $\mathcal{L}_{\text{Mult}}$ over the set $\mathcal{N}^{(L)}(s)$ of nodes at most $L$ steps from $s$:
\begin{equation}
    \mathcal{L}_{\text{Mult}} = \frac{1}{|\mathcal{G}_{\text{test}}|}\sum_{G \in \mathcal{G}_{\text{test}}}\sum_{v \in \mathcal{N}^{(L)}(s)} \abs{1 - \frac{y_v}{\Phi_v(G)}}
    \label{eq:test-loss}
\end{equation}
where $y_v$ is the true label for each reachable node. The MinAggGNN reaches a train loss of $\mathcal{L}_{\MSE} = 0.0001$ and a test loss of $\mathcal{L}_{\text{Mult}} = 0.0578$ (\cref{fig:shortest_path_loss}).

\subsection{The Bellman-Ford Circuit}
\label{subsec:bellman-ford-circuit}

\begin{figure}
    \centering
    \includegraphics[width=.49\linewidth]{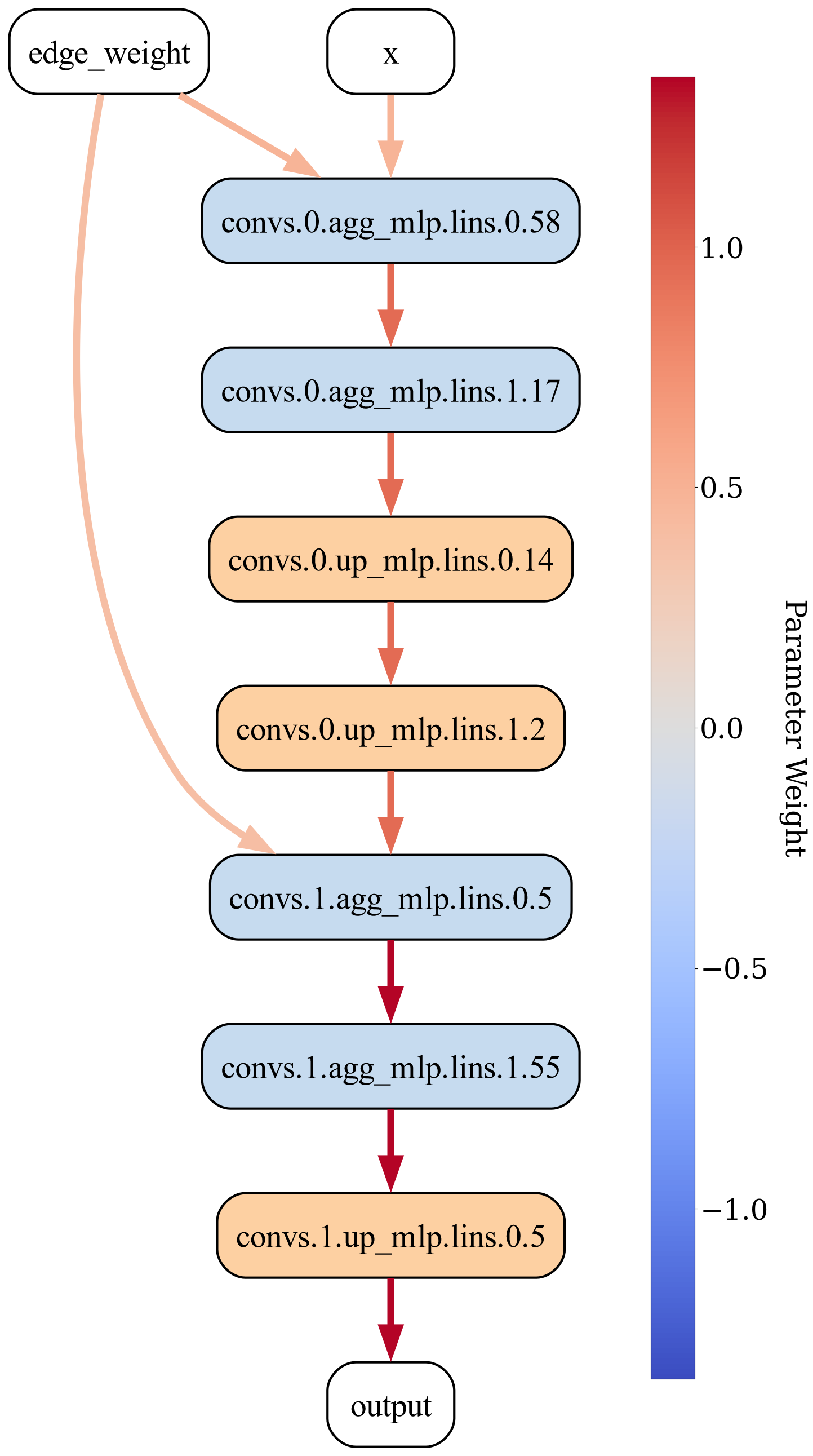}
    \includegraphics[width=.49\linewidth]{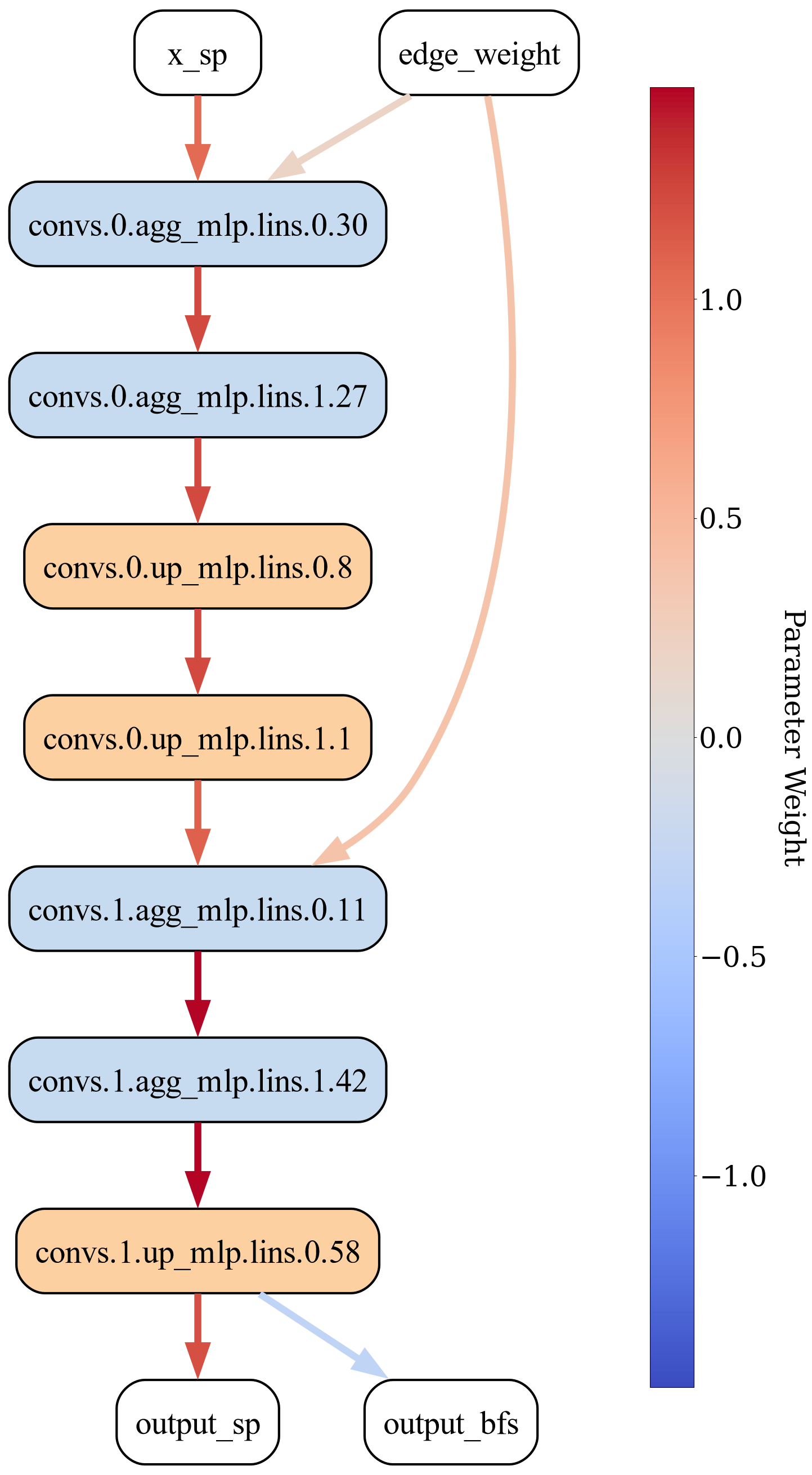}
    \caption{Identified circuit in the Bellman-Ford MinAggGNN (left) and parallel Bellman-Ford and BFS MinAggGNN (right). Nodes are individual MinAggGNN neurons. Input and output neurons are colored white, $f_{\Agg}$ neurons are colored blue, and $f_{\Up}$ neurons are colored orange. Circuit edges are colored by corresponding model weights.}
    \label{fig:bellman-ford-circuit}
\end{figure}

For circuit discovery, we corrupt the input data by setting every edge weight to zero and flipping the input features:
\begin{equation}
    {x_v^{\text{SP}}}' = \begin{cases}
    B & v \text{ is the source} \\
    0 & \text{otherwise}.
    \end{cases}
\end{equation}
Using this corruption with MSE, we use each scoring method in \cref{sec:method} to identify a circuit in the 18240-edge computation graph. All methods except \textsc{WeightGrad} are able to identify the the theoretically optimal circuit of 10 parameters (\cref{fig:bellman-ford-circuit}, left) predicted by~\citet{nerem2025graphneuralnetworksextrapolate}. (A comparison of scoring methods is given in \cref{appdx:bf-score-comparison}.)

The circuit is sufficient to replicate the model's Bellman-Ford implementation: taken as a subnetwork of the full model, the circuit achieves $\mathcal{L}_{\text{Mult}} = 0.0545$ on the test set, matching the high performance of the full model. (In fact, the circuit achieves a slightly lower loss overall.) It is also necessary: when the circuit is removed, the resulting loss is $\mathcal{L}_{\text{Mult}} = 21870.6836$. In this example, MINAR successfully recovers the minimal circuit that captures the behavior of the full model, validating our approach.

\subsection{Circuit Pruning and Grokking}
\label{subsec:circuit-grokking}

One motivation for mechanistic interpretability study is the \emph{grokking} phenomenon, where a model exhibits strong generalizability well after its training statistics appear to converge~\cite{power2022grokkinggeneralizationoverfittingsmall}. Several mechanisms have been put forth to explain why grokking occurs, including regularization~\cite{power2022grokkinggeneralizationoverfittingsmall, nanda2023progress, zhong2023clock, li2025interpretationpredictbehaviorunseen}, representation learning~\cite{liu2022towards, liu2023omnigrok, zhong2023clock, mallinar2025emergence}, emergence of information-theoretic phenomena~\cite{clauw2024informationtheoretic, demoss2025complexity}, or other properties of training dynamics~\cite{barak2022hidden, kumar2024grokking}.

\begin{figure}
    \centering
    \includegraphics[width=.95\linewidth]{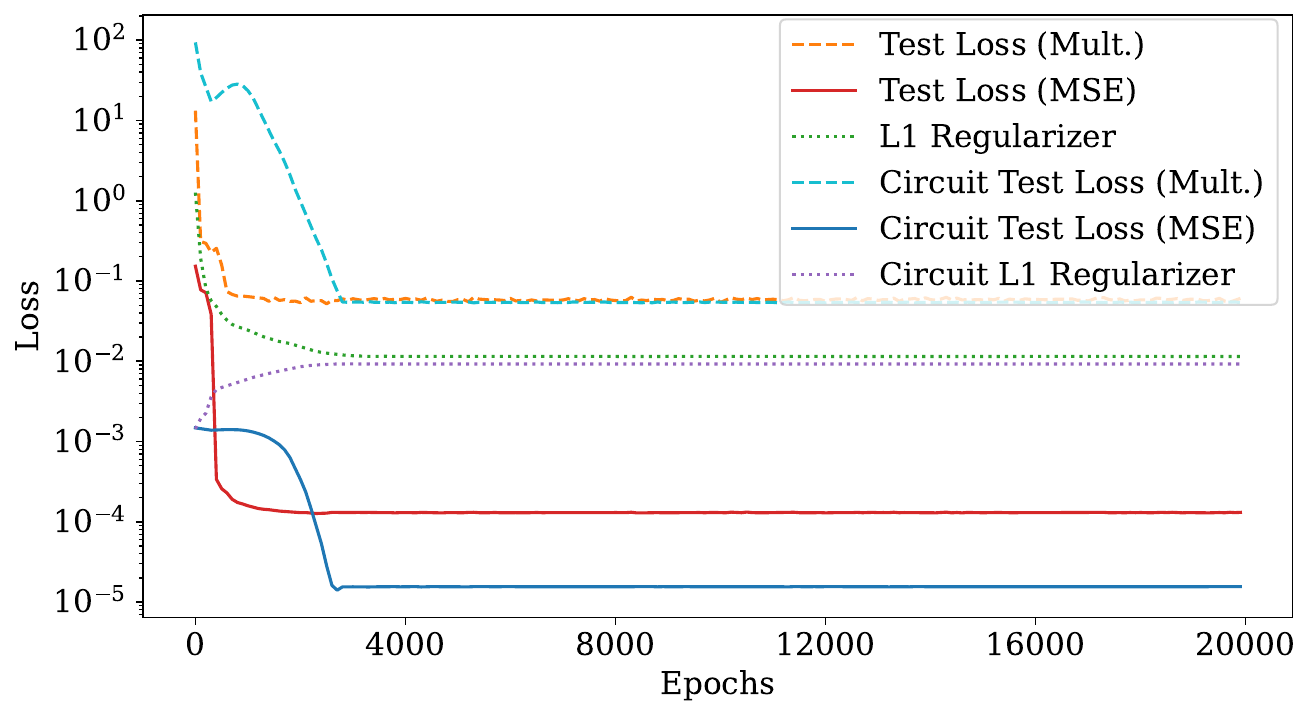}
    \caption{Multiplicative test loss, MSE test loss, and $L_1$ regularization terms for the Bellman-Ford network and circuit.}
    \label{fig:bellman_ford_circuit_grokking}
\end{figure}

In our work, we observe that the Bellman-Ford network generalizes almost immediately. In fact, the generalization occurs almost simultaneously with the convergence of the training loss.
Curiously, however, the parameters that eventually form the Bellman-Ford circuit generalize later than the full model, as shown in \cref{fig:bellman_ford_circuit_grokking}. While the full model performs well on the test set after only about $1000$ epochs, the final circuit does not generalize until about epoch $3000$ as the $L_1$ regularization stabilizes.
Examining this phase of training, MINAR reveals that a generalizable circuit does appear as early as epoch 1000, but it is significantly larger than the final circuit. As the model continues to train with $L_1$ regularization, we can see the minimal sufficient circuit shrink as edges are pruned.
\cref{appdx:sparsification} shows circuits extracted from this intermediate stage of training. A more striking example is given in \cref{appdx:no-self-loops}, where a concise circuit emerges early, but a \emph{minimal} circuit does not appear until around 18000 epochs.

Our findings support observations by \cite{elhage2022toymodelssuperposition, nanda2023progress, zhong2023clock, li2025interpretationpredictbehaviorunseen} that once a neural circuit emerges during training, regularization can prune extraneous parameters from the model. For example, \citet{elhage2022toymodelssuperposition} show a similar relationship between model capacity and neural representations: as model size grows, \emph{polysemantic} neurons (which are responsible for mutiple concepts) split into \emph{monosemantic} neurons (which are responsible for only one concept). Expanding model capacity further allows for redundant representations, where a single concept may be represented by multiple neurons.

However, the nature of pruned parameters that we observe differs from other examples: \citet{li2025interpretationpredictbehaviorunseen} show the emergence of ``vestigial circuits'' alongside generalizable circuits that implement non-generalizable heuristics or memorize parts of the training data. These vestigial circuits must then be pruned later in training by weight decay or regularization. In contrast, we observe that the extra parameters early in training are actually a necessary part of the neural circuit. That is, rather than forming and pruning early heuristic circuits, we observe that regularization serves to refine a generalizing model into a more concise circuit.

This delayed pruning also has implications for techniques like model compression~\cite{Caruana2006compression, han2016compressing} or distillation~\cite{hinton2015distillingknowledgeneuralnetwork, salimans2022progressive, panigrahi2025progressive}, which seek to emulate the behavior of a large performant model using a much smaller model. The smaller model may be pruned directly from the larger model, or trained using outputs from the larger model in a student-teacher manner.
Our results show that strong test performance---even out-of-distribution performance---may not be enough to characterize a desirable teacher model. Rather, we show that continued training with $L_1$ regularization past model generalization can produce a model with sparser, more efficient circuits, in turn producing a model more amenable to compression.

\subsection{Shortcut Learning: Using Bellman-Ford for BFS}
\label{subsec:bf-bfs-parallel}

We extend the Bellman-Ford example by introducing a second task: predicting reachability from a source node by BFS. \citet{velickovic2020Neural} note that Bellman-Ford and BFS traverse a graph in the same manner, and show that training these tasks in parallel can benefit performance on both. Here, MINAR reveals surprising behavior in this two-task setup: the model learns a shortcut for BFS based entirely on the shortest path distance.

\begin{figure}
    \centering
    \includegraphics[width=.99\linewidth]{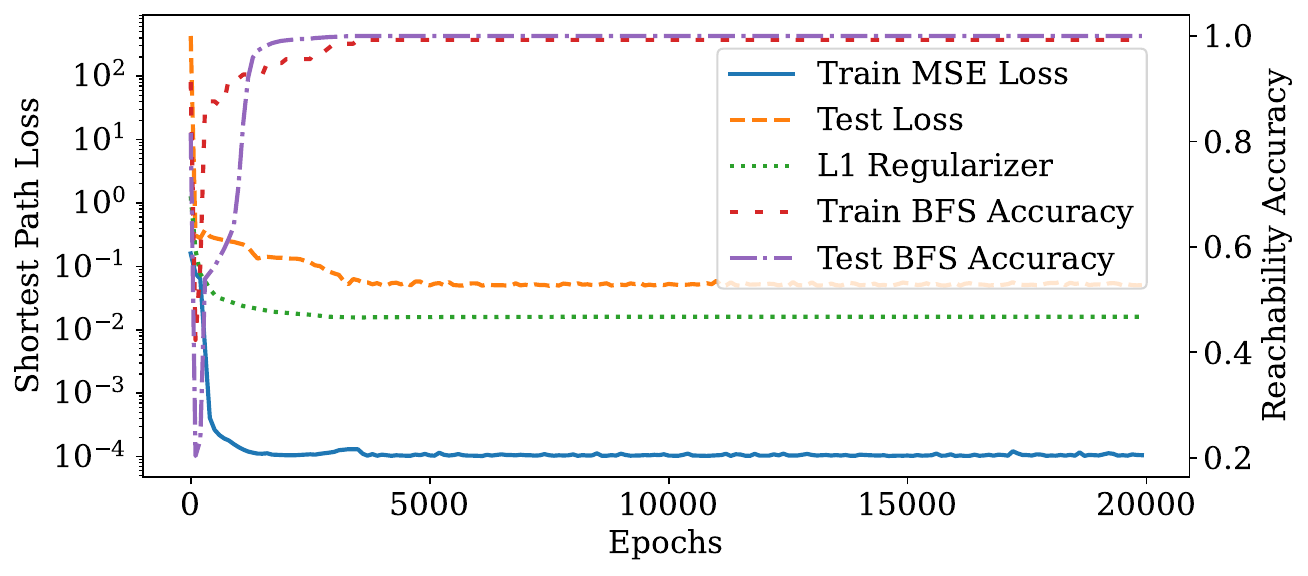}
    \caption{MSE training loss, multiplicative test loss, reachability accuracy, and $L_1$ regularization term for parallel Bellman-Ford and BFS MinAggGNN.}
    \label{fig:parallel_loss}
\end{figure}

Both tasks share the same training set, but with additional node features and labels to support the reachability task. As with Bellman-Ford, the model's two layers correspond to two steps of BFS. In addition to the Bellman-Ford feature \eqref{eq:sp-encoding} each node is given an initial feature based for BFS from the same source source $s$:
\begin{equation}
    x_v^{\text{BFS}} = \begin{cases}
    1 & v \text{ is the source} \\
    0 & \text{otherwise}.
    \end{cases}
    \label{eq:bfs-encoding}
\end{equation}
Thus the parallel GNN receives as input the concatenation $(x_v^{\text{SP}}, x_v^{\text{BFS}})$. For the test set, we additionally include a number of balanced tree graphs to provide more graphs with unreachable nodes.
We train simultaneously, using MSE for the shortest path task and binary cross-entropy (BCE) for the reachability task (more details are given in \cref{appdx:training}).
\Cref{fig:parallel_loss} shows the loss and accuracy curves of the model during training, with a test loss of 0.0519 on shortest path and a test accuracy of 1.00 on reachability.

We again use EAP-IG with $m=20$ steps to identify the circuit .
Again, we corrupt the input data by setting every edge weight to zero and flipping the input features:
\begin{equation}
    {x_v^{\text{BFS}}}' = \begin{cases}
    0 & v \text{ is the source} \\
    1 & \text{otherwise}.
    \end{cases}
\end{equation}
We identify a circuit (\Cref{fig:bellman-ford-circuit}, right) with 11 edges (out of 18432) that achieves $\mathcal{L}_{\text{Mult}} = 0.0499$ on shortest paths and a reachability accuracy of 0.9831, showing that the circuit is sufficient. It is also necessary: removing it results in $\mathcal{L}_{\text{Mult}} = 58038.1680$ and a test accuracy of 0.8198 (the accuracy when the model classifies every node as reachable).

Upon inspection, it is clear that the circuit does not perform BFS independently---it is almost identical to the Bellman-Ford circuit. Manual inspection of the parameters represented in the BFS circuit reveals that the BFS output has
\begin{equation}
    \begin{split}
        &\texttt{output\_bfs} =\\
        &\phantom{-}(-.2927)\texttt{convs.1.up\_mlp.lins.0.58} + 2.6463.
    \end{split}
\end{equation}
That is, the model has learned a shortcut: multiply the internal representation of the shortest path distance by a negative weight and compare against the bias term. While concerns about shortcut learning often center around poor out-of-distribution generalization~\cite{geirhos2020shortcut}, MINAR exposes shortcut learning which is invisible even to a large out-of-distribution test set.

%% file: sec/6-salsa-clrs.tex
\section{Case Study: SALSA-CLRS}
\label{sec:salsa-clrs}

\begin{table*}
    \centering
    \begin{tabular}{|c|c|c|c|c|c|c|c|}
    \hline
        Algorithm & BFS & DFS & Dijkstra & Prim's MST & Bellman-Ford & Articulation Points & Bridges \\
        \hline
         Val Acc. (\%)  & 99.90 & 70.73 & 93.37 & 81.15 & 93.75 & 94.63 & 78.68 \\
         Test Acc. (\%) & 98.25 & 36.42 & 81.50 & 48.18 & 82.55 & 97.42 & 88.22
         \\ \hline
    \end{tabular}
    \caption{Validation and test accuracy of maximum-aggregated GINE on SALSA-CLRS.}
    \label{tab:salsa-clrs-val}
\end{table*}
\begin{figure*}
    \centering
    \includegraphics[width=\linewidth]{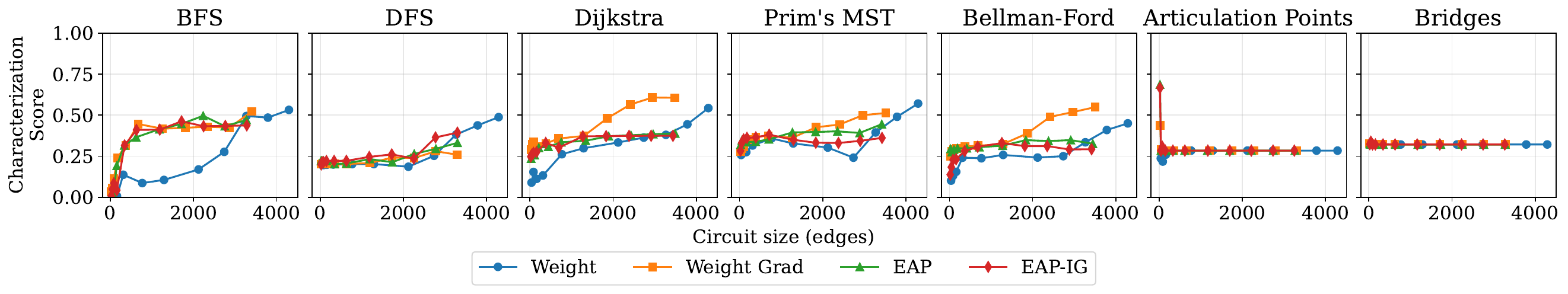}
    \caption{Characterization score of circuits computed using Weight, \textsc{WeightGrad}, EAP, and EAP-IG ($m=20$) for SALSA-CLRS GNN for $K \in \{10, 25, 50, 100, 250, 500, 1000, 1500, 2000, 2500, 3000\}$.}
    \label{fig:score-comparison}
\end{figure*} 
\begin{figure}
    \centering
    \includegraphics[width=\linewidth]{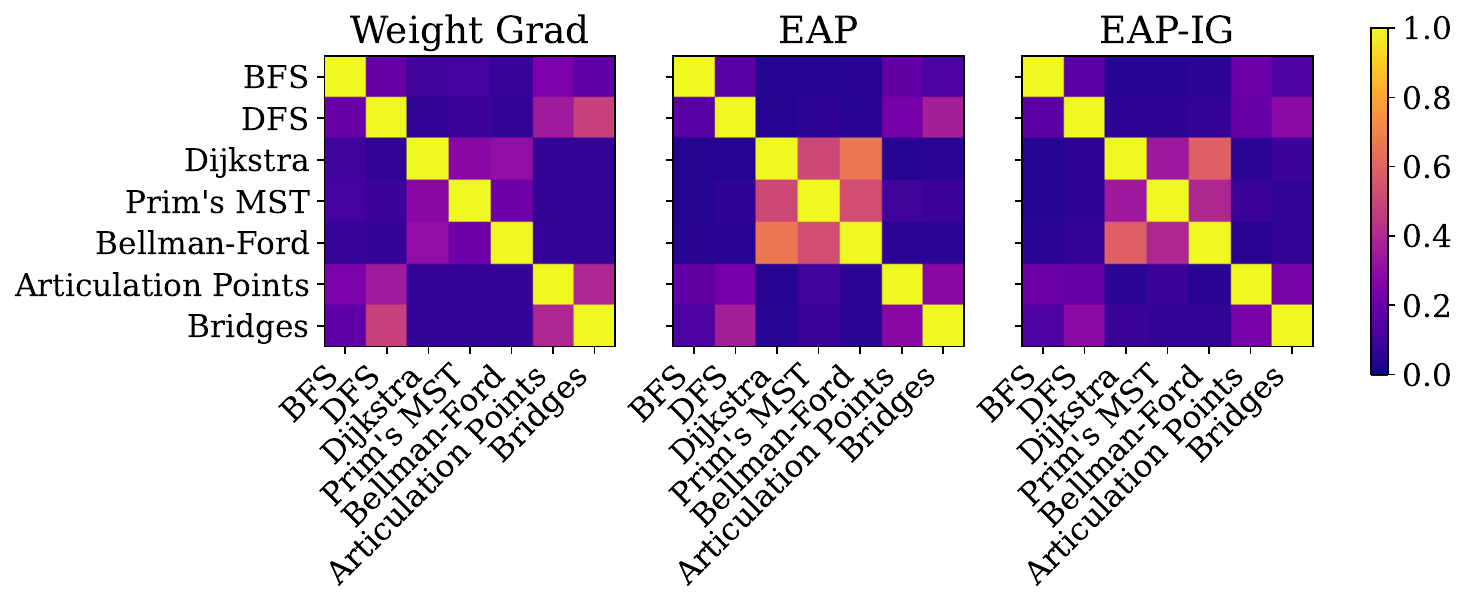}
    \caption{Weighted Jaccard Similarity of task-specific circuits with \textsc{WeightGrad}, EAP, EAP-IG at $K=500$.}
    \label{fig:circuit-overlap}
\end{figure}

In this section we apply circuit discovery to a larger collection of graph-based reasoning tasks taken from the CLRS-30 benchmark~\cite{velickovic2022clrs}, providing a more challenging case study. Specifically, we use the PyTorch Geometric implementations of breadth-first search (BFS), depth-first search (DFS), Dijkstra's shortest path algorithm, and Prim's minimum spanning tree (MST) algorithm from SALSA-CLRS~\cite{minder2023salsaclrs}. We also include implementations of Bellman-Ford, Articulation Points, and Bridges from CLRS-30 in the SALSA-CLRS framework. These seven tasks represent the subset of CLRS tasks which can be represented on sparse graphs.
\footnote{The remaining CLRS-30 tasks require complete graphs to represent problem instances, as CLRS-30's hints allow arbitrary node connections. We restrict ourselves to tasks where hints are restricted to edges which already exist in the graph. For further discussion, see \cref{appdx:clrs-data}.}

\subsection{Architecture and Training}

Following the Encode-Process-Decode setup described in~\cite{velickovic2022clrs, minder2023salsaclrs}, each task has its own set of linear encoders and decoders for node and edge features.
The processor is a graph neural network $\Phi$ shared across all algorithmic tasks, which is applied recurrently to the encoded features. We use a 2-layer Graph Isomorphism Network with Edge Features (GINE)~\cite{hu2020Strategies}, modified from the Graph Isomorphism Network in~\cite{xu2018how}. Each layer is given by  
\begin{equation}
    \Phi_v^{(\ell+1)} = f^{(\ell)} \left(
    \Phi_v^{(\ell)} + \max_{u \in \mathcal{N}(v)} \ReLU(\Phi_u^{(\ell)} + W^{(\ell)}_e e_{u,v}) \right),
    \label{eq:gine-conv}
\end{equation}
where $f^{(\ell)}$ is a two-layer MLP and $W^{(\ell)}_e$ is a linear map which maps the edge feature $e_{u,v}$ to the same dimension as the node embedding space. Training details are provided in \cref{appdx:training}.
The trained model does not generalize perfectly, as some tasks are known to be challenging~\cite{velickovic2022clrs, minder2023salsaclrs}. It is also not as sparse as the Bellman-Ford models in \cref{sec:bellman-ford}. Despite the lack of optimality in the trained model, we show that MINAR can still extract faithful circuits that offer insight into the model's behavior.

\subsection{Circuit Discovery and Analysis}

To perform circuit discovery on the full network, we begin by creating corrupted datasets for each task. We do so by zero-ablating every input feature on unseen graphs drawn from the validation distribution. This includes node positions on all graphs; source nodes for BFS, DFS, Dijkstra's, Prim's, and Bellman-Ford; and edge weights on Dijkstra's, Prim's, and Bellman-Ford. Because the processor is responsible for the algorithmic reasoning steps, we focus on identifying circuits within the processor GNN using the same loss that was used in training. A more detailed discussion can be found in \cref{appdx:clrs-data}.

We use \cref{alg:circuit_discovery} to identify circuits in the processor's computation graph of 99712 edges using edge weights, \textsc{WeightGrad}, EAP, and EAP-IG ($m=20$), and evaluate the fidelity of each circuit with respect to the full model $\Phi$. The \emph{positive fidelity} $\text{Fid}^+$ measures the \emph{necessity} of a circuit $C$ while the \emph{negative fidelity} $\text{Fid}^-$ measures the \emph{sufficiency}: denote by $\Phi^{\setminus C}$ the model without the circuit and by $\Phi^{C}$ the model using only the circuit weights:
\begin{equation}
    \begin{split}
        \text{Fid}^+(C) &= 1 - \frac{1}{|V|}\sum_{v \in G}\mathbf{1}\{\Phi_v(G) \neq \Phi^{\setminus C}_v(G)\} \\
        \text{Fid}^-(C) &= 1 - \frac{1}{|V|}\sum_{v \in G}\mathbf{1}\{\Phi_v(G) \neq \Phi^{C}_v(G)\}.
    \end{split}
    \label{eq:fidelity}
\end{equation}
We combine these metrics using the \emph{characterization score} from~\cite{amara2022graphframex}. The characterization score is the harmonic mean of $\text{Fid}^+(C)$ and $(1 - \text{Fid}^-(C))$:
\begin{equation}
    \textsc{Char}(C) = \left[\frac{1}{2\text{Fid}^+(C)} + \frac{1}{2(1 - \text{Fid}^-(C))}\right]^{-1}.
    \label{eq:characterization-score}
\end{equation}
\cref{fig:score-comparison} shows the characterization score for values of $K$ up to 3000.
(Individual fidelity scores and accuracies of circuits and their ablations are given in \cref{appdx:additional-results-clrs}.)
We can see that \textsc{WeightGrad}, EAP, and EAP-IG consistently outpace the weight baseline in finding explanatory circuits for most tasks. Surprisingly, despite its simplicity, \textsc{WeightGrad} performs quite well, frequently exceeding EAP and EAP-IG. These results show that even on a denser, more challenging network, MINAR successfully identifies faithful circuits using only a fraction of the computation graph.

As in \cref{subsec:bf-bfs-parallel}, it is natural to see if the SALSA-CLRS model also leverages the same circuit components to perform related tasks. For example, Dijkstra's algorithm, Prim's algorithm, and Bellman-Ford all rely on edge weights, while the standard approaches for Articulation points and Bridges both rely on DFS. We measure circuit overlap by the weighted Jaccard similarity between the circuit edge sets, where edges are weighted by the absolute value of their corresponding model weights. \cref{fig:circuit-overlap} displays the overlap of circuits identified with \textsc{WeightGrad}, EAP, and EAP-IG at $K=500$. (Conservatively, at $K=500$ each circuit is $< 1\%$ of the computation graph. We do not plot the overlap of circuits found with Weight, as it is invariant to the task and therefore all the circuits are identical.) Additional overlap comparisons for different scoring methods can be found in \cref{appdx:additional-results-clrs}. We can see across scoring methods that, as expected, the circuits for Dijkstra's algorithm, Prim's algorithm, and Bellman-Ford have significant overlap. We also see strong overlap between DFS, Articulation Points, and Bridges. Interestingly, EAP and EAP-IG circuits show stronger edge weight overlap while \textsc{WeightGrad} identifies the overlap in DFS-based tasks first. This circuit overlapping phenomenon provides a potential explanation for transfer learning observations like~\cite{velickovic2020Neural}, where training on similar tasks in parallel helps GNNs perform on each individual task.

%% file: sec/7-conclusion.tex
\section{Conclusion}
\label{sec:conclusion}

In this work, we introduce MINAR, the first (to our knowledge) attempt to use automated circuit discovery in the GNN setting. Through two case studies, we demonstrate how MINAR enables a detailed analysis of GNNs trained to solve algorithmic tasks, with implications for the role of circuit formation and pruning in grokking and model compression. We also show, using MINAR, how GNNs trained to perform multiple algorithmic tasks in parallel share circuit components for similar tasks---in the most extreme case exposing severe shortcut learning. We hope that by bringing the emerging field of mechanistic interpretability to the study of neural algorithmic reasoning, MINAR provides a useful tool to study NAR models in detail.

%% file: sec/ack.tex
\section*{Acknowledgements}
We thank Robert R. Nerem and Samantha Chen for discussions regarding the Bellman-Ford MinAggGNN.
This work was conducted under the Laboratory Directed Research and Development Program at PNNL, a multi-program national laboratory operated by Battelle for the U.S. Department of Energy under contract DE-AC05-76RL01830.
It was also partially supported by NSF CCF-2112665, CCF-2217058, CCF-2403452, and CCF-2310411.

%% file: sec/impact.tex
\section*{Impact Statement}

This work advances the study of neural algorithmic reasoning by enabling mechanistic interpretability of GNN models. While interpretability is of societal concern, this work is of a fundamental nature and hence we do not feel that any specific impact must be highlighted here.

%% file: sec/appdx/additional-bf-results.tex
\section{Additional Results for Bellman-Ford}

This appendix contains additional results from experiments in \cref{sec:bellman-ford} (\cref{appdx:sparsification}), as well as relevant plots from runs with different training data (\cref{appdx:no-self-loops}) and seeds (\cref{appdx:seeds}).

\subsection{Score Comparison for Bellman-Ford}
\label{appdx:bf-score-comparison}

\begin{figure}[h]
    \centering
    \includegraphics[width=\linewidth]{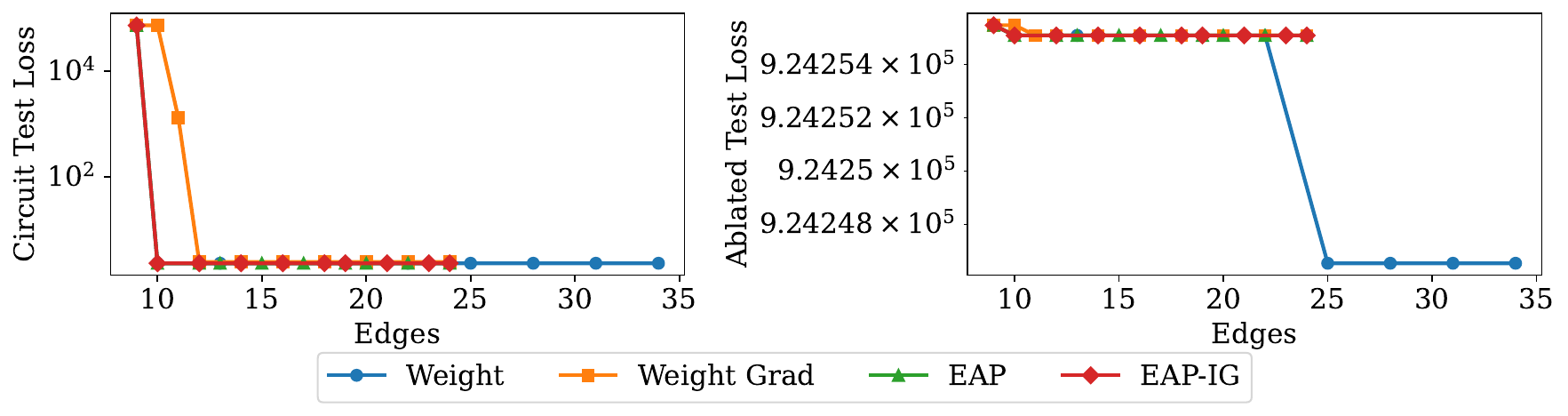}
    \caption{Circuit loss (left) and ablated circuit loss (right) for circuits found using Weight, \textsc{WeightGrad}, EAP, and EAP-IG ($m=20$) with $K = 1, 2, \dots, 10$ for the Bellman-Ford network.}
    \label{fig:bf-score-comparison}
\end{figure}

\begin{figure}[h]
    \centering
    \includegraphics[width=0.35\linewidth]{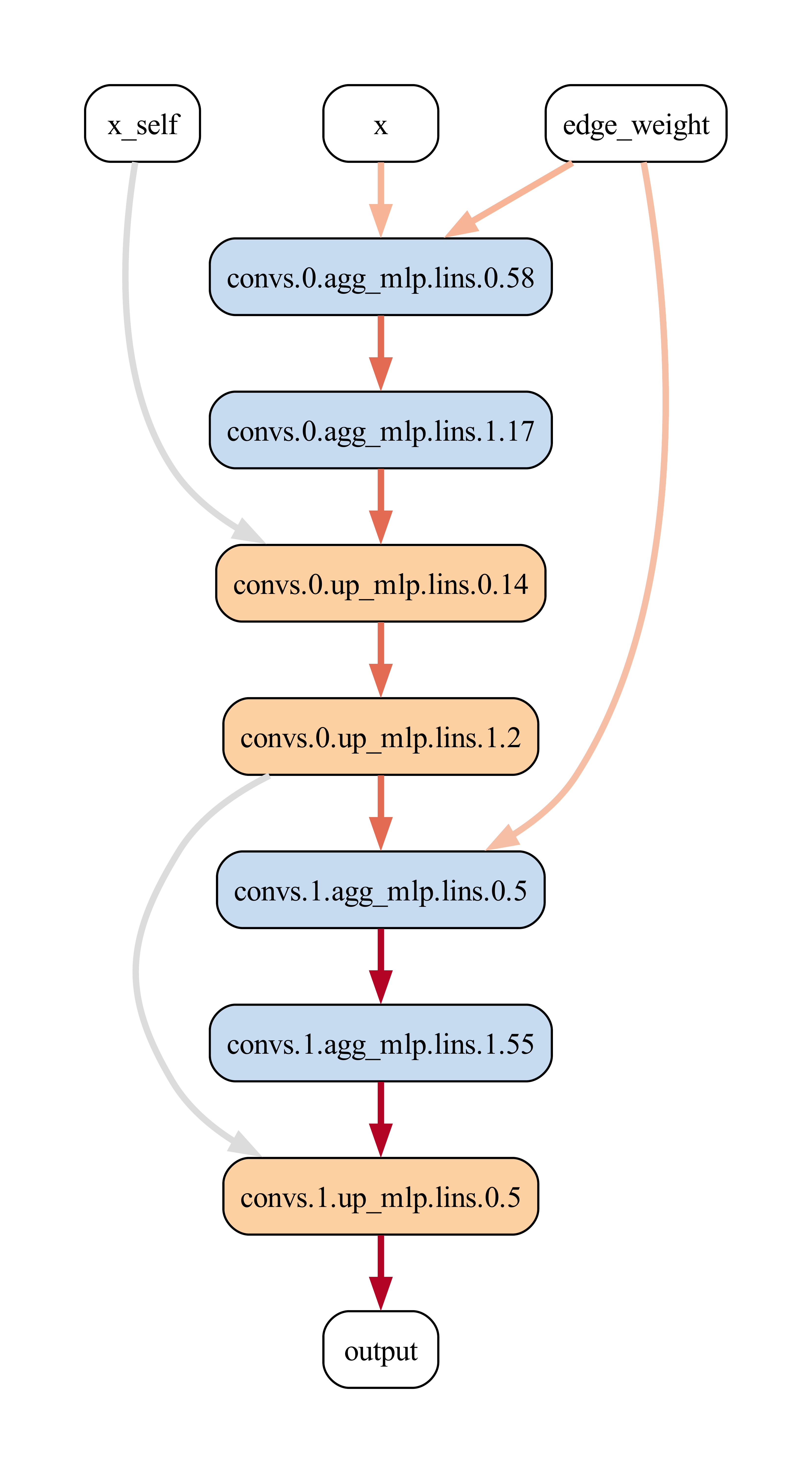}
    \caption{Best Bellman-Ford circuit found by \textsc{WeightGrad}.}
    \label{fig:bf-weight-grad-circuit}
\end{figure}

We compare in \cref{fig:bf-score-comparison} different scoring methods for the Bellman-Ford GNN in \cref{subsec:bellman-ford-circuit}. Only \textsc{WeightGrad} fails to find the optimal 10-edge circuit, instead requiring 12 edges (\cref{fig:bf-weight-grad-circuit}).

\subsection{Circuit Sparsification}
\label{appdx:sparsification}

Here we demonstrate how the circuit in \cref{subsec:bellman-ford-circuit} (\cref{fig:bellman-ford-circuit}) emerges during training by extracting circuits from earlier epochs. We identify sufficient circuits at epochs 1000 and 2000. The final circuit appears by epoch 3000.
This mirrors observations in LLMs where features may not align with individual neurons, but limiting capacity (either architecturally or in our case through regularization) discourages instances where a single feature is represented by multiple neurons~\cite{elhage2022toymodelssuperposition}. (In contrast, expanding model capacity has been observed in to encourage so-called \emph{polysemantic} neurons to split into several \emph{monosemantic} neurons in LLMs.)

\begin{figure}[ht]
    \centering
    \includegraphics[width=\textwidth]{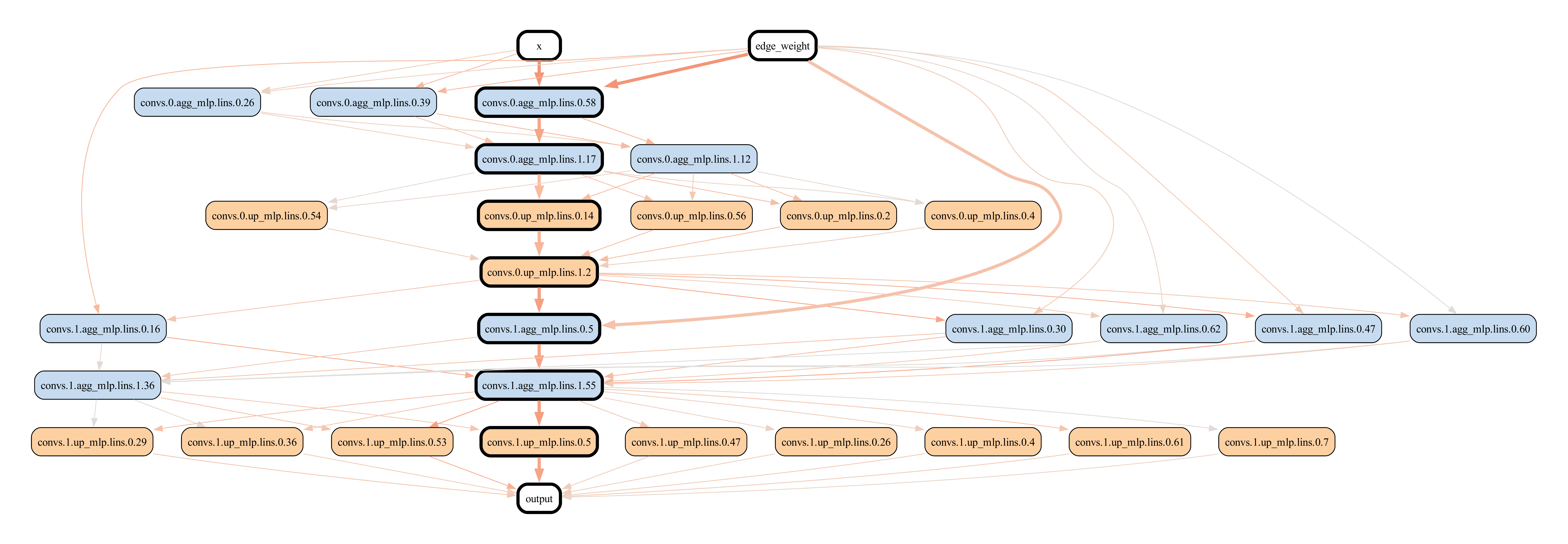}
    \caption{Bellman-Ford circuit at epoch 1000. The circuit contains 73 edges and achieves $\mathcal{L}_{\text{Mult}} = 0.0578$. Elements which are part of the final circuit are bolded.}
    \label{fig:bf-circuit-epoch-1000}
\end{figure}

\begin{figure}[ht]
    \centering
    \includegraphics[width=.85\textwidth]{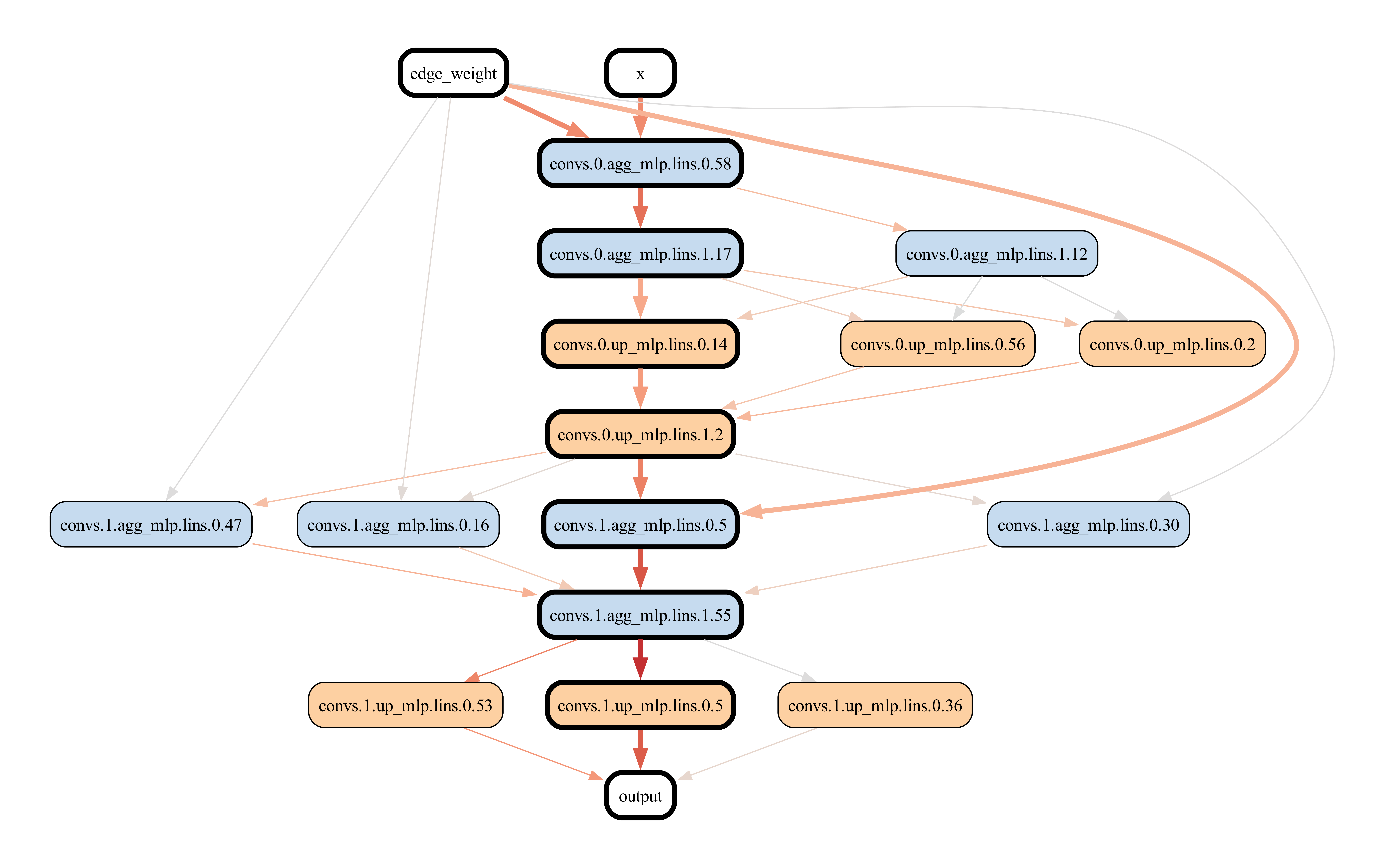}
    \caption{Bellman-Ford circuit at epoch 2000. The circuit 31 edges and achieves $\mathcal{L}_{\text{Mult}} = 0.0565$. Elements which are part of the final circuit are bolded.}
    \label{fig:bf-circuit-epoch-2000}
\end{figure}

\newpage
\subsection{Bellman-Ford without Self-Loops}
\label{appdx:no-self-loops}

A previous version of this work contained experiments from \cref{sec:bellman-ford} in which the training and testing graphs did not have self-loops. As a result, while the minimum aggregation in \cref{eq:minagg-gnn} returned the optimal neighbor embedding for each node, it did not include the node's own embedding. Hence the update MLP $f_{\text{Up}}$ became responsible for comparing between a node's current shortest path distance and that of its best neighbor. While self-loops have been added in \cref{sec:bellman-ford}, we briefly study the behavior of the model trained without self-loops to highlight how MINAR reveals the emergence of this ability to compute the minimum between two elements.

\begin{figure}[hp]
    \centering
    \includegraphics[width=0.5\linewidth]{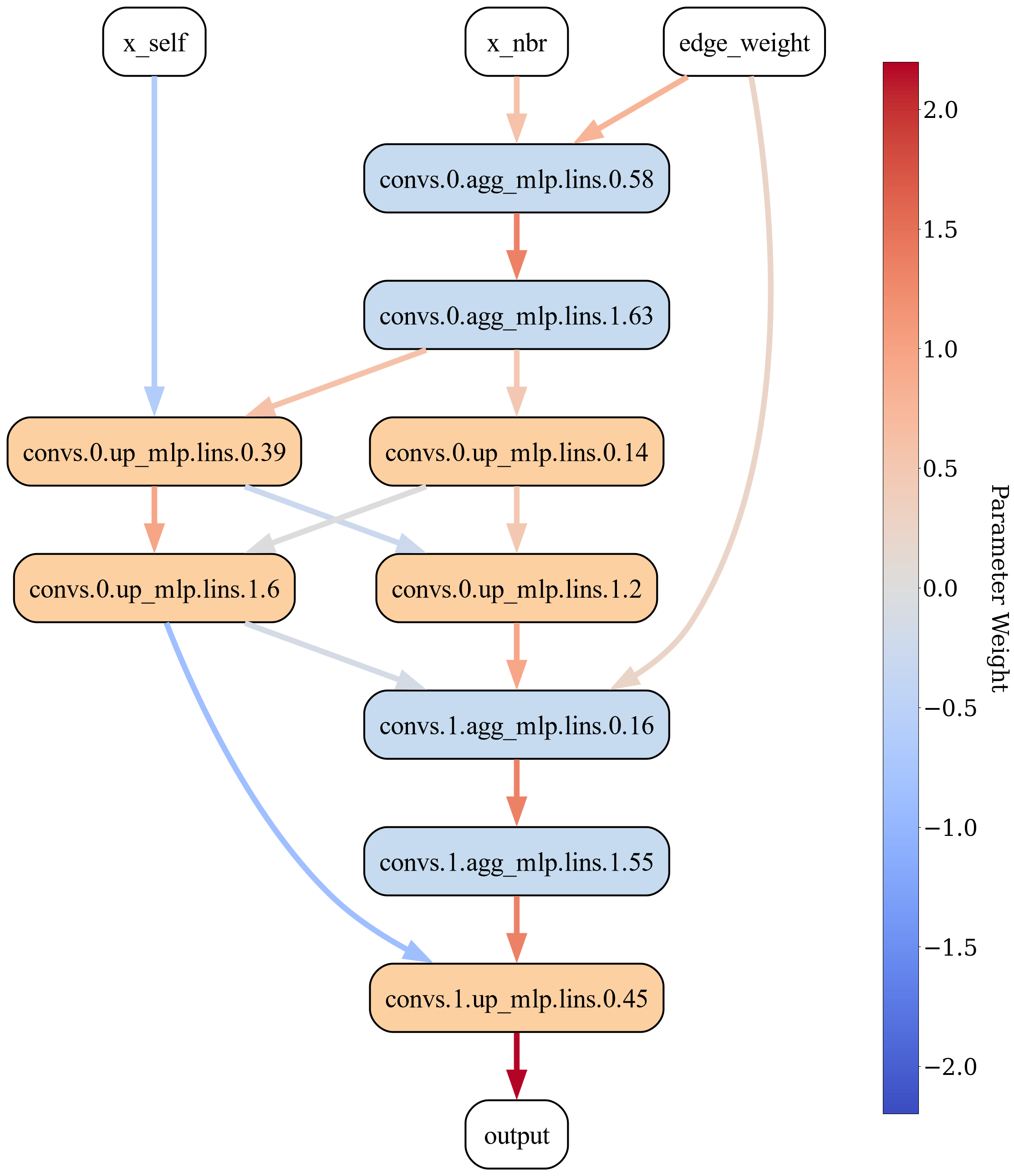}
    \caption{Bellman-Ford circuit from MinAggGNN trained without self-loops. Contains 17 of 18,240 edges. $\mathcal{L}_{\text{Mult}} = 0.0599$.}
    \label{fig:no-self-loop-circuit}
\end{figure}

The MinAggGNN is trained on the same data as \cref{sec:bellman-ford}, achieving an MSE Loss of $\mathcal{L}_{\text{MSE}} = 0.0002$ and $\mathcal{L}_{\text{Mult}} = 0.0642$. The circuit (\cref{fig:no-self-loop-circuit}), consisting of 17 edges and discovered using EAP-IG ($m=20$) achieves a test loss of $\mathcal{L}_{\text{Mult}} = 0.0599$.

\begin{figure}[hp]
    \centering
    \includegraphics[width=0.5\linewidth]{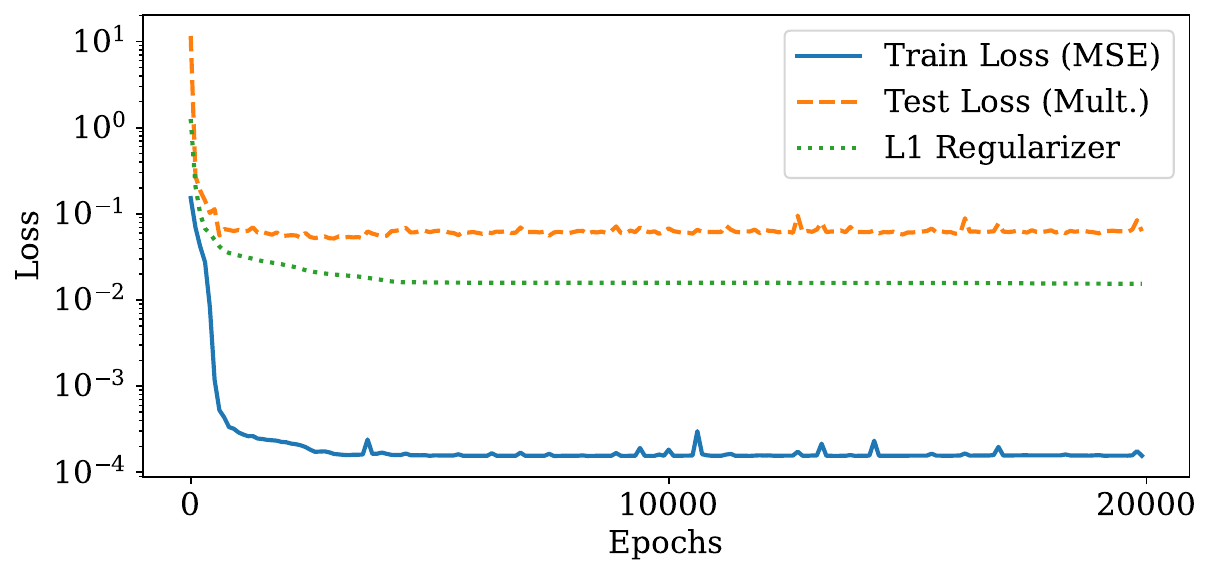}
    \caption{Train loss, test loss, and $L_1$ regularization term for Bellman-Ford MinAggGNN trained without self-loops.}
    \label{fig:no-self-loop-loss}
\end{figure}

\newpage
Here, we wish to highlight the nodes for the first update MLP (beginning with \texttt{convs.0.up\_mlp}). Manually inspecting the weights, we see that this portion of the circuit implements the minimum operation between the initial node embedding $x_{\text{self}}$ and the best possible update among its neighbors in the first step, which we will denote $x_{\text{nbr}}^*$.
\begin{equation}
    \texttt{convs.0.up\_mlp.lins.1.6}
        = \begin{bmatrix}
            0 & \phantom{-}0.9620
        \end{bmatrix}
        \ReLU\left(
            \begin{bmatrix}
                0.4921 & 0 \\
                0.5905 & -0.6102
            \end{bmatrix}
            \begin{bmatrix}
                x_{\text{nbr}}^* \\ x_{\text{self}}
            \end{bmatrix}
        \right)
\end{equation}
If $0.5905 x_{\text{nbr}}^* - 0.6102 x_{\text{self}} > 0$ (i.e. if roughly $x_{\text{nbr}}^* > x_{\text{self}}$), then this becomes
\begin{equation}
    \begin{bmatrix}
        0.5681 & -0.5870
    \end{bmatrix}
    \begin{bmatrix}
        x_{\text{nbr}}^* \\ x_{\text{self}}
    \end{bmatrix}
    \propto x_{\text{nbr}}^* - x_{\text{self}}.
\end{equation}
Otherwise, if $0.5905 x_{\text{nbr}}^* - 0.6102 x_{\text{self}} < 0$ (i.e. if roughly $x_{\text{nbr}}^* < x_{\text{self}}$), then this becomes 0.

Notice in \cref{fig:no-self-loop-circuit} the primary input to the second step of Bellman-Ford comes from \texttt{convs.0.up\_mlp.lins.1.2}. If $x_{\text{nbr}}^* < x_{\text{self}}$, then this is the optimal representation for Bellman-Ford. However, if $x_{\text{nbr}}^* > x_{\text{self}}$, the computation will incorrectly carry forward with $x_{\text{nbr}}^*$ instead of $x_{\text{self}}$. In this case, by storing a value proportional to $x_{\text{nbr}}^* - x_{\text{self}}$, the neuron \texttt{convs.0.up\_mlp.lins.1.6} can be rescaled and subtracted from the input to the final update MLP, serving as a correction factor.

The upshot of this calculation is that the update MLP $f_{\Up}^{(0)}$ essentially learns to implement the minimum operation by
\begin{equation}
    \min\{x_{\text{nbr}}^*, x_{\text{self}}\} = x_{\text{nbr}}^* - \ReLU(x_{\text{nbr}}^* - x_{\text{self}}).
\end{equation}

\begin{figure}[p]
    \centering
    \includegraphics[width=\linewidth]{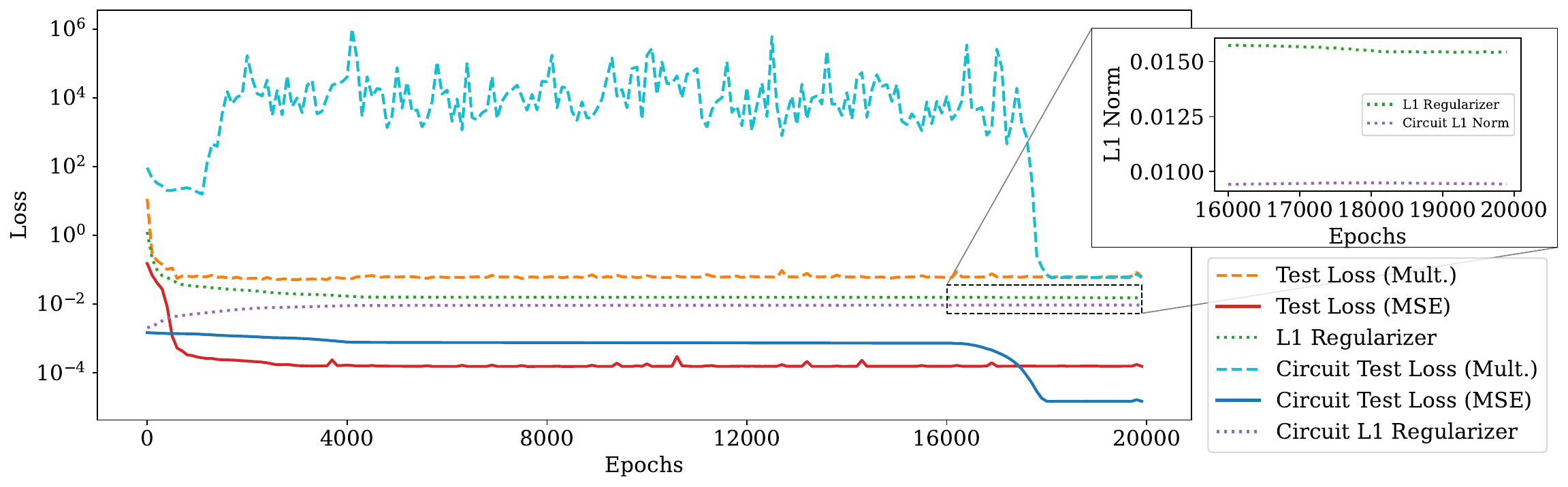}
    \caption{Multiplicative test loss, MSE test loss, and $L_1$ regularization terms for the full Bellman-Ford network without self-loops and the Bellman-Ford circuit without self-loops. Inset shows a small inflection point in $L_1$ norms.}
    \label{fig:no-self-loop-grokking}
\end{figure}

\begin{figure}[p]
    \centering
    \includegraphics[width=0.5\linewidth]{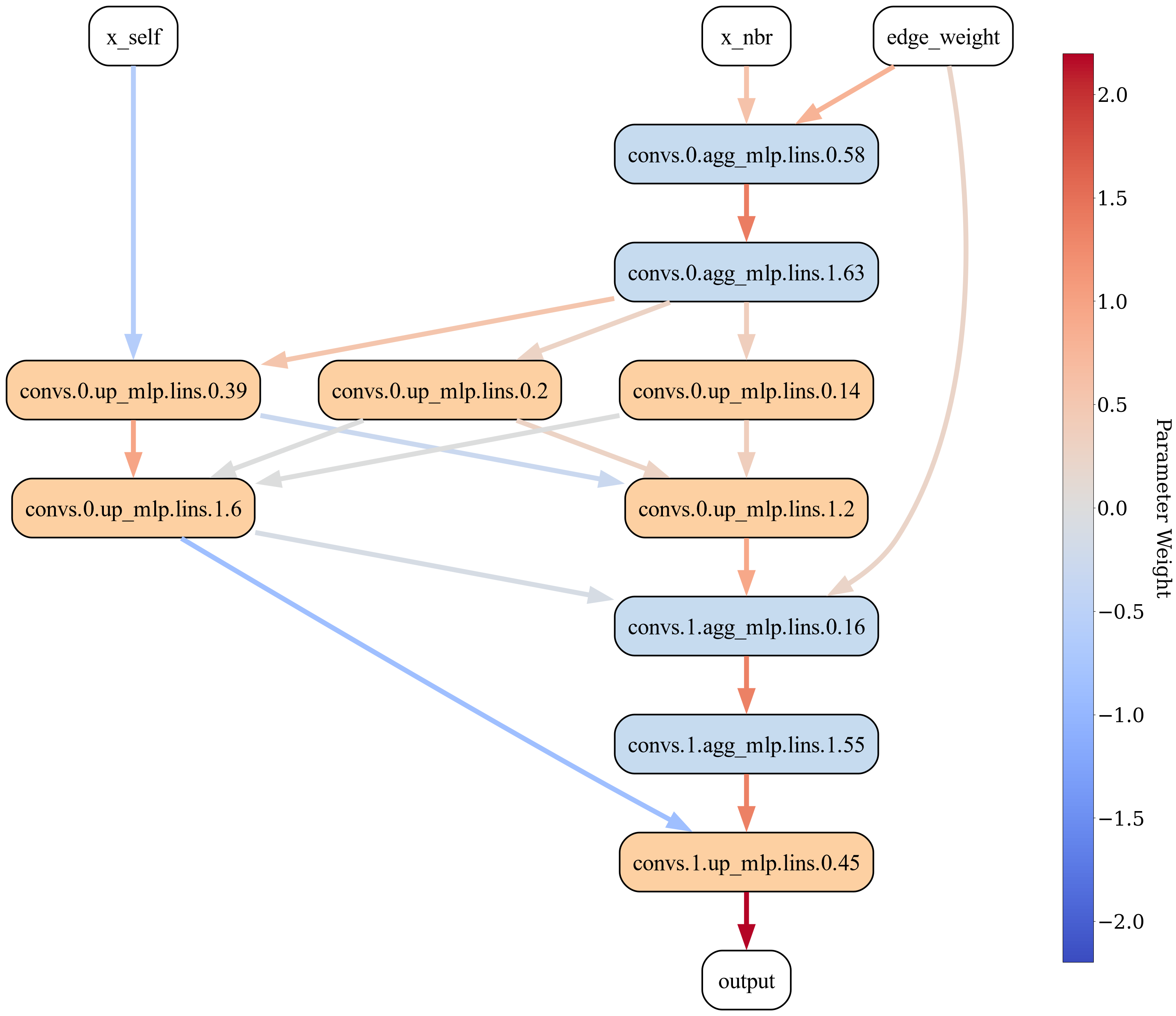}
    \caption{Bellman-Ford circuit extracted from epoch 10000 of MinAggGNN training without self-loops. Contains 20 of 18,240 edges.}
    \label{fig:no-self-loop-circuit-mid}
\end{figure}

We also highlight the generalization behavior of the circuit in \cref{fig:no-self-loop-grokking}. While the full model achieves strong generalization performance early, as in \cref{sec:bellman-ford}, the final circuit does not generalize until much later (around epoch 18000). Examining a circuit extracted from an earlier checkpoint (epoch 10000) in \cref{fig:no-self-loop-circuit-mid}, we can see that the connections which are pruned away at this later inflection point (namely, the connections to and from \texttt{convs.0.up\_mlp.lins.0.2}) are partially responsible for this same minimum operation in the first update MLP. As in \cref{appdx:sparsification}, we see $L_1$ regularization encourage the model to align its internal representations with the neuron basis to achieve a sparser solution.

\newpage
\subsection{Results from different runs}
\label{appdx:seeds}

Here we show relevant plots from different experimental trials.

\subsubsection{Bellman-Ford}

\begin{figure}[htp]
    \centering
    \begin{subfigure}{0.24\textwidth}
        \includegraphics[width=\textwidth]{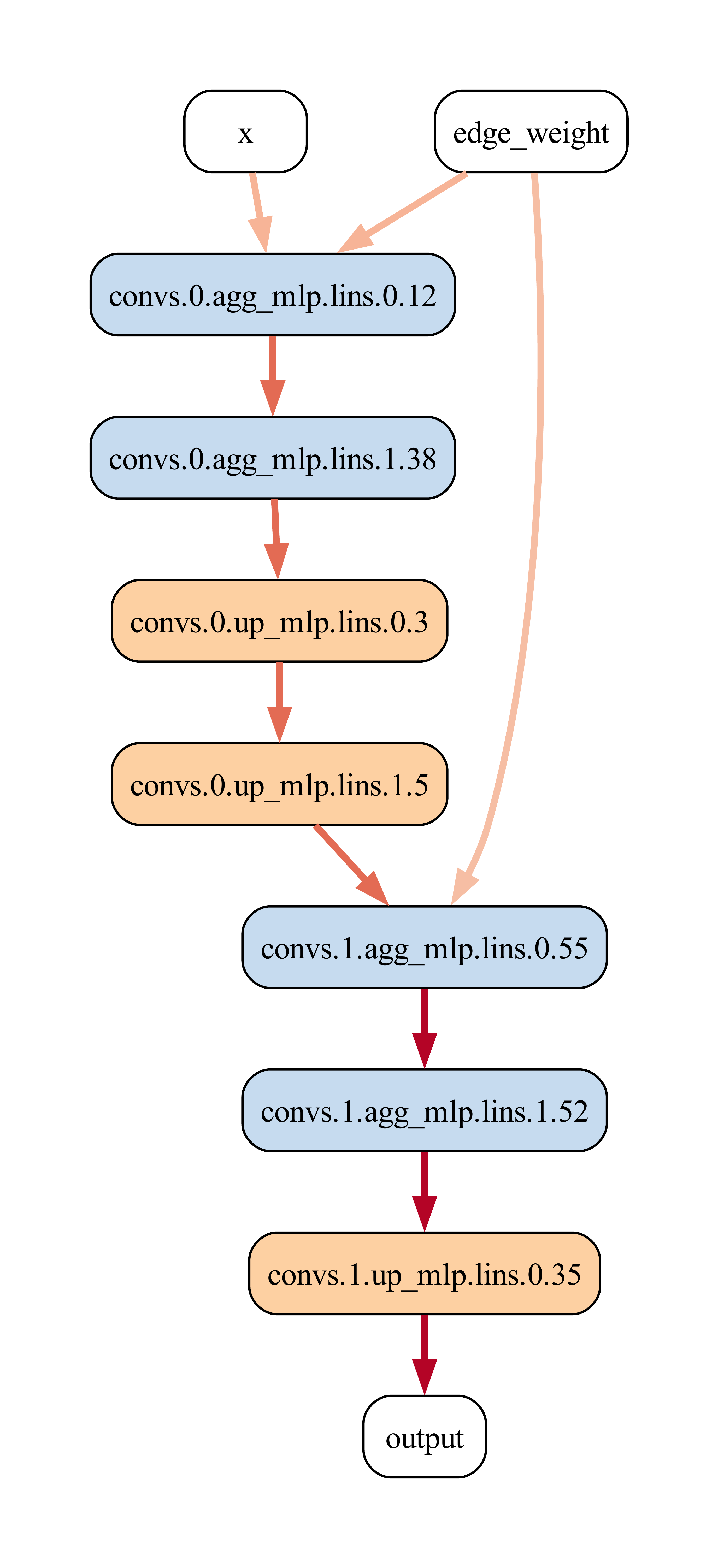}
        \label{subfig:bf-circuit-1}
        \caption{Seed 1. $\mathcal{L}_{\text{Mult}} = 0.0543$}
    \end{subfigure}
    \begin{subfigure}{0.24\textwidth}
        \includegraphics[width=\textwidth]{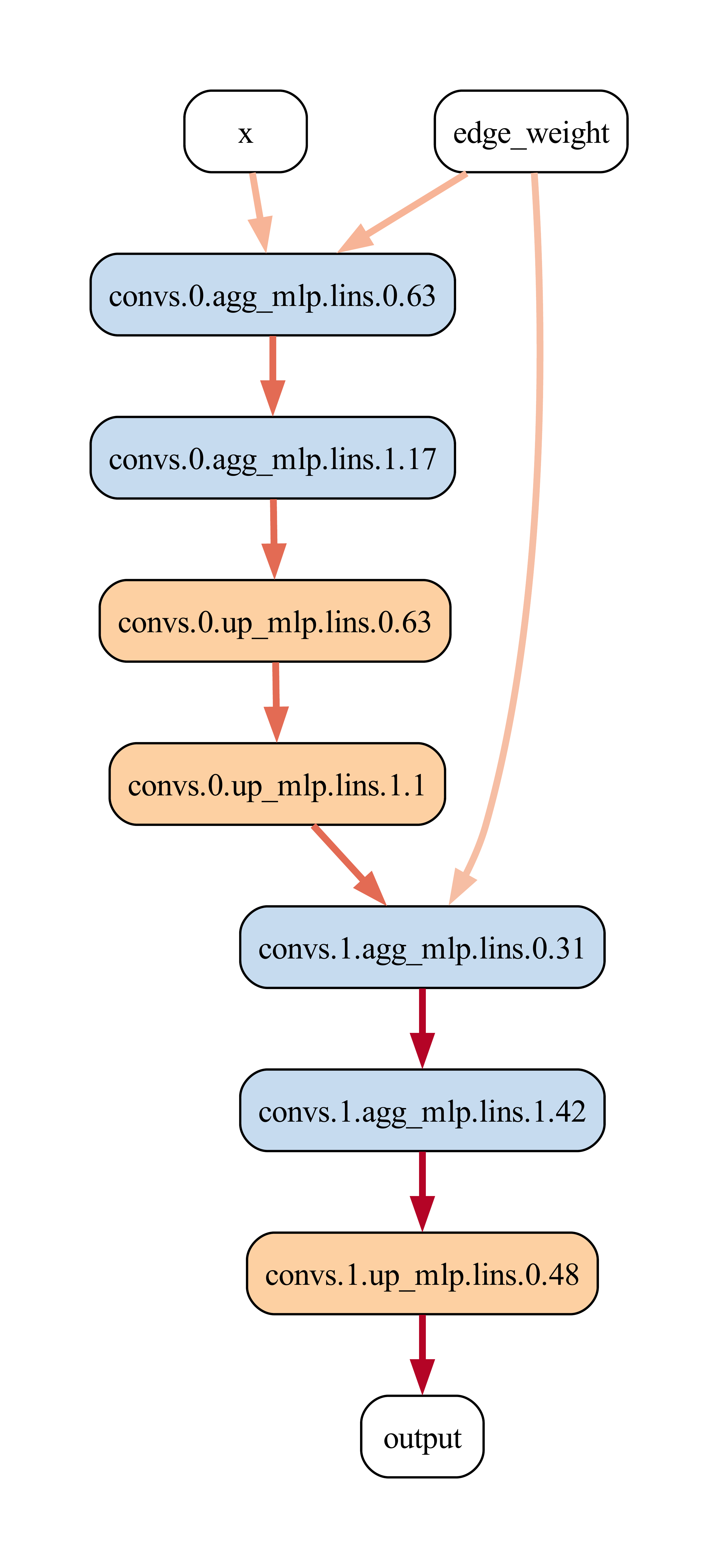}
        \label{subfig:bf-circuit-2}
        \caption{Seed 2. $\mathcal{L}_{\text{Mult}} = 0.0540$}
    \end{subfigure}
    \begin{subfigure}{0.24\textwidth}
        \includegraphics[width=\textwidth]{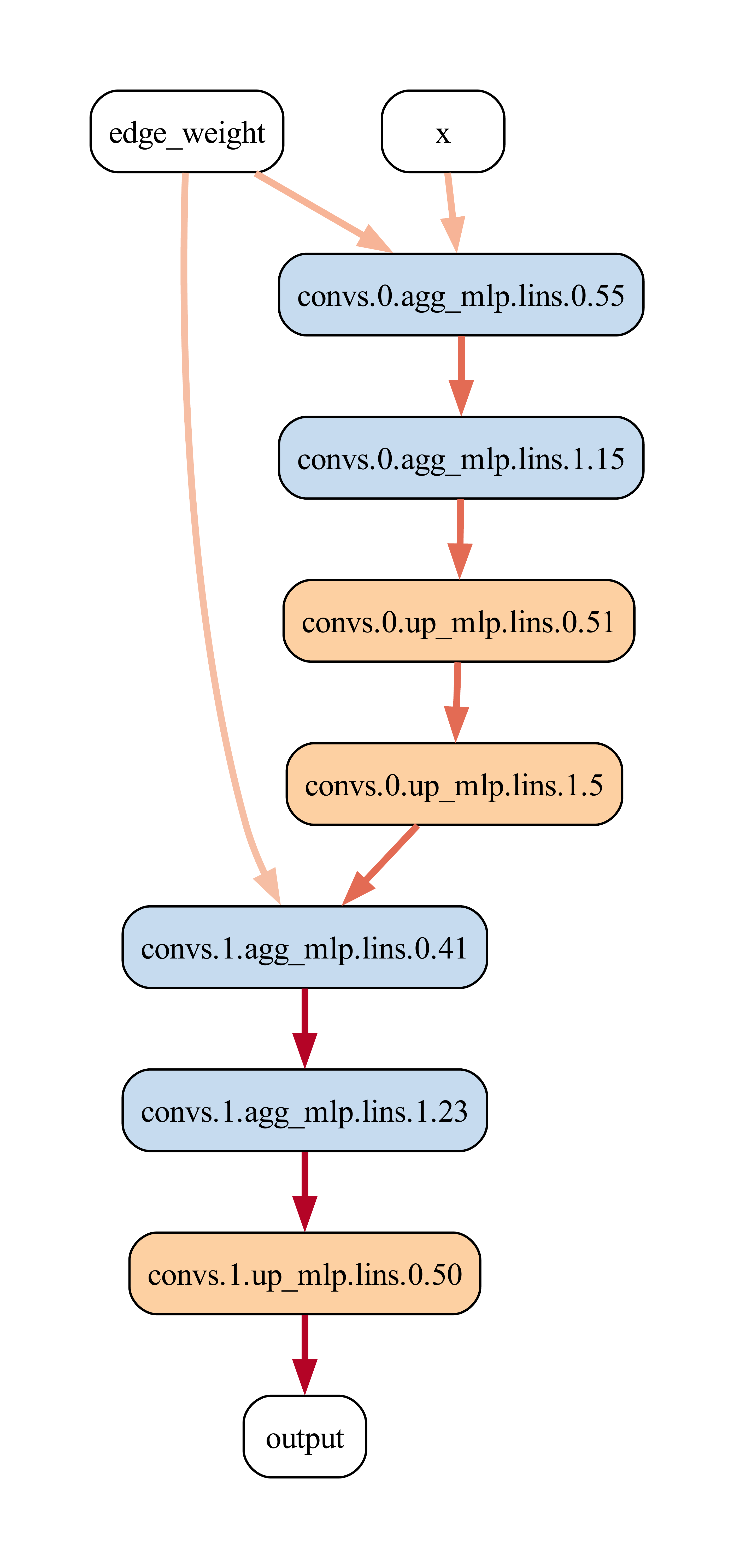}
        \label{subfig:bf-circuit-3}
        \caption{Seed 3. $\mathcal{L}_{\text{Mult}} = $}
    \end{subfigure}
    \begin{subfigure}{0.24\textwidth}
        \includegraphics[width=\textwidth]{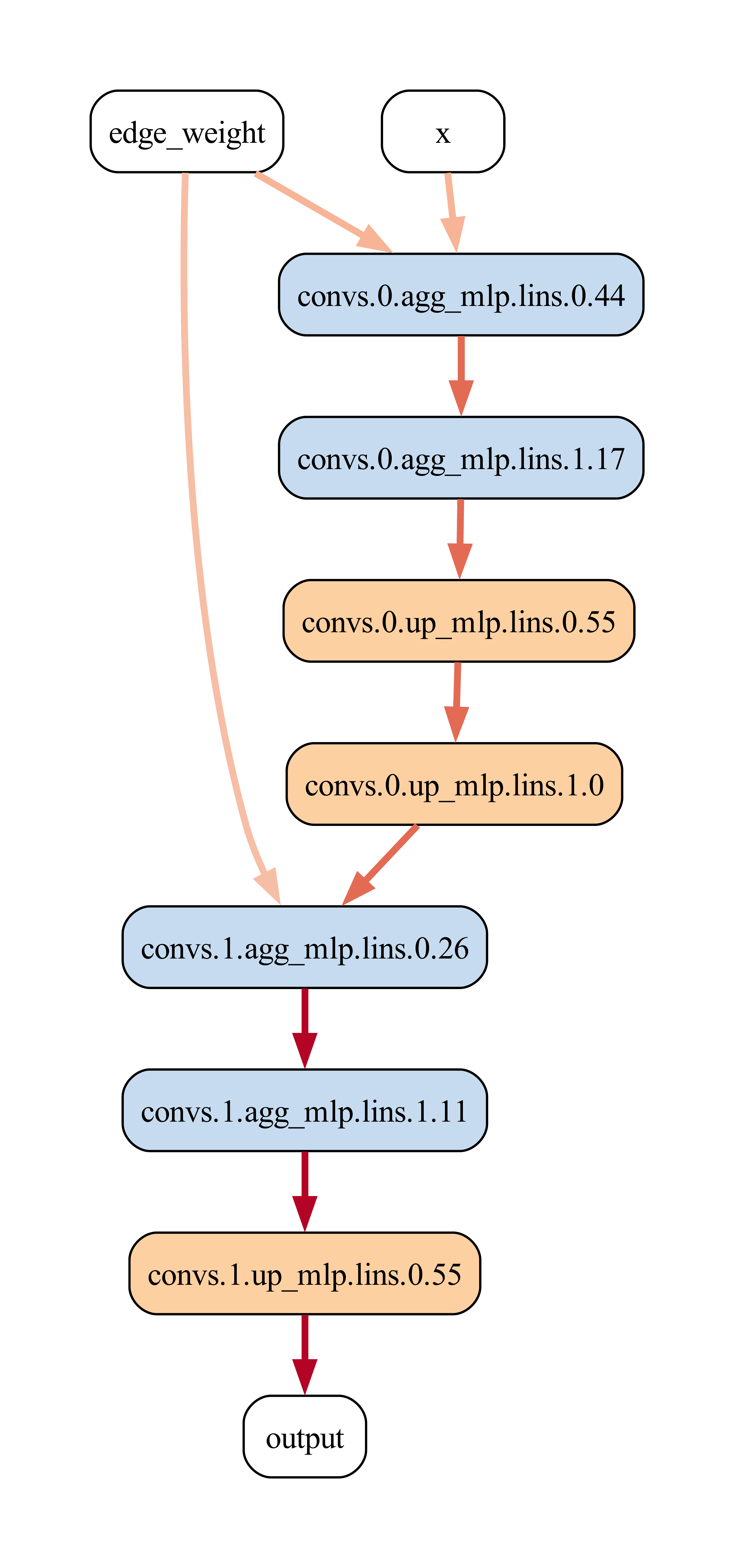}
        \label{subfig:bf-circuit-4}
        \caption{Seed 4. $\mathcal{L}_{\text{Mult}} = $}
    \end{subfigure}
    \label{fig:bf-circuit-runs}
    \caption{Bellman-Ford circuits across four additional initializations.}
\end{figure}

We show circuits extracted across different initializations for experiments in \cref{subsec:bellman-ford-circuit} in \cref{fig:bf-circuit-runs}. Each circuit is functionally identical to the mainline run, but contain different neurons.

\begin{figure}[htp]
    \centering
    \begin{subfigure}{0.49\textwidth}
        \includegraphics[width=\textwidth]{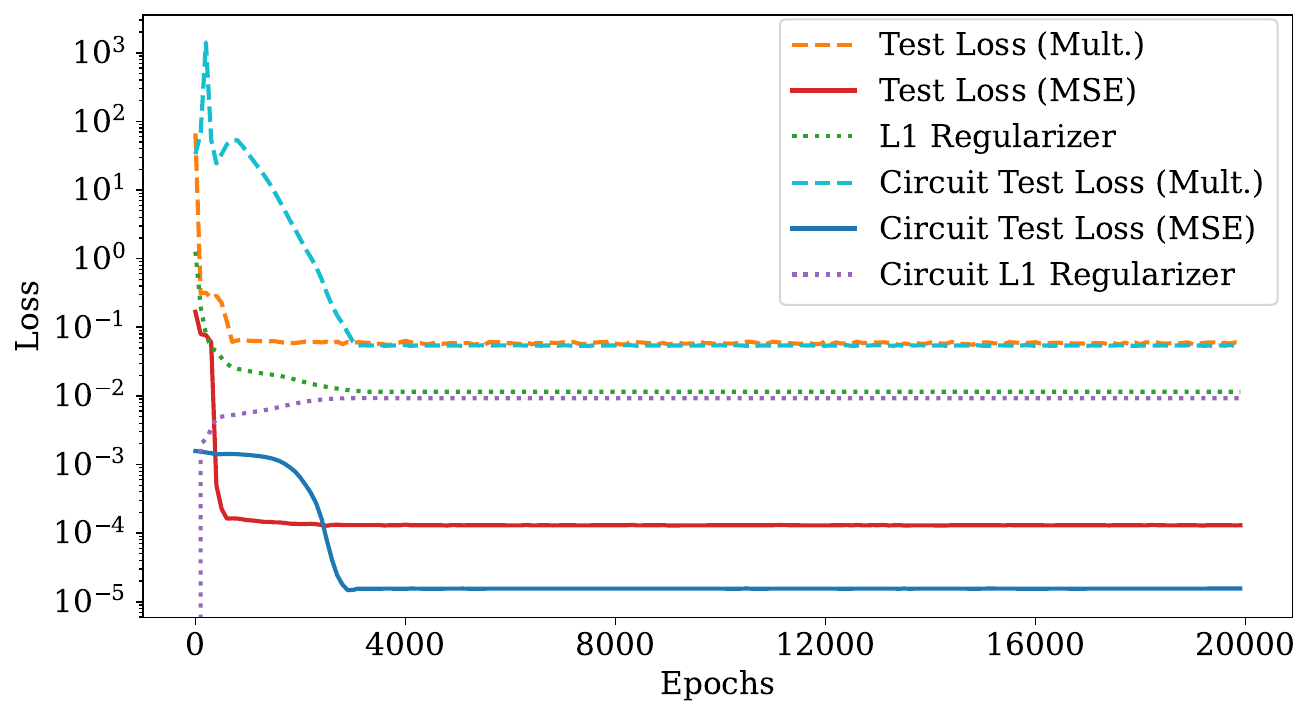}
        \label{subfig:bf-grok-1}
        \caption{Seed 1}
    \end{subfigure}
    \begin{subfigure}{0.49\textwidth}
        \includegraphics[width=\textwidth]{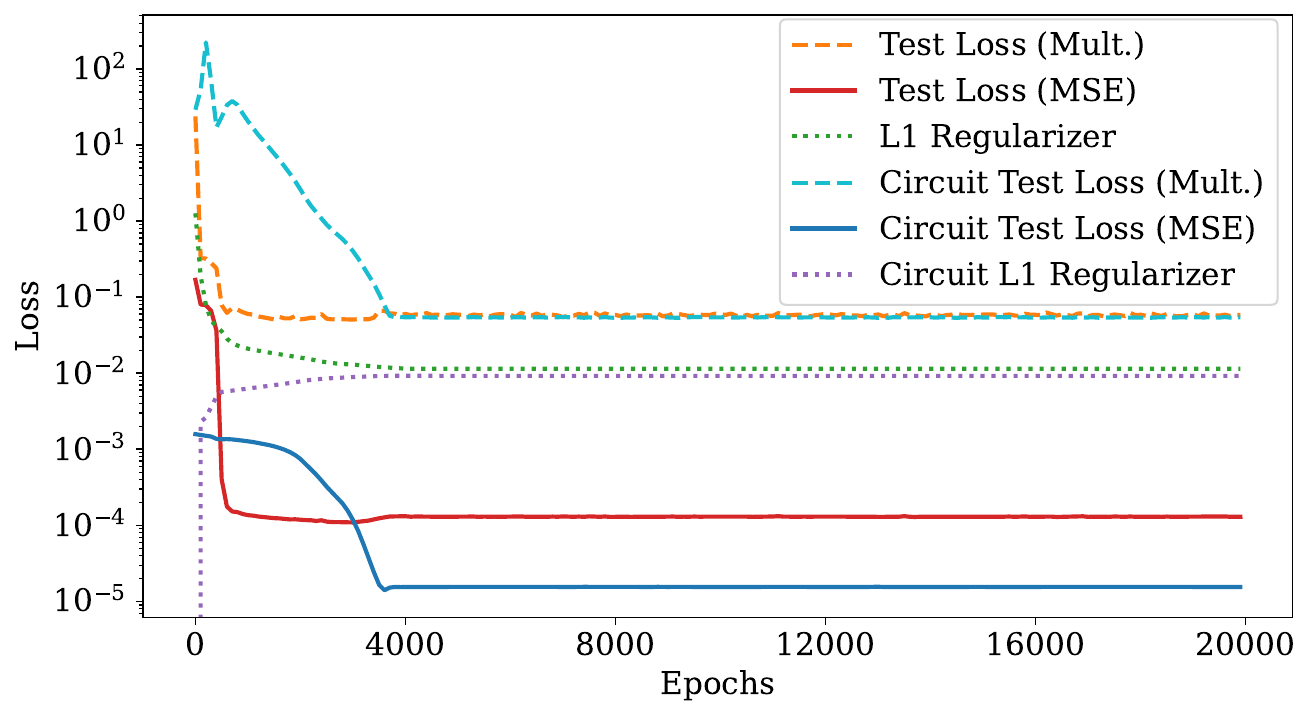}
        \label{subfig:bf-grok-2}
        \caption{Seed 2}
    \end{subfigure}
    \begin{subfigure}{0.49\textwidth}
        \includegraphics[width=\textwidth]{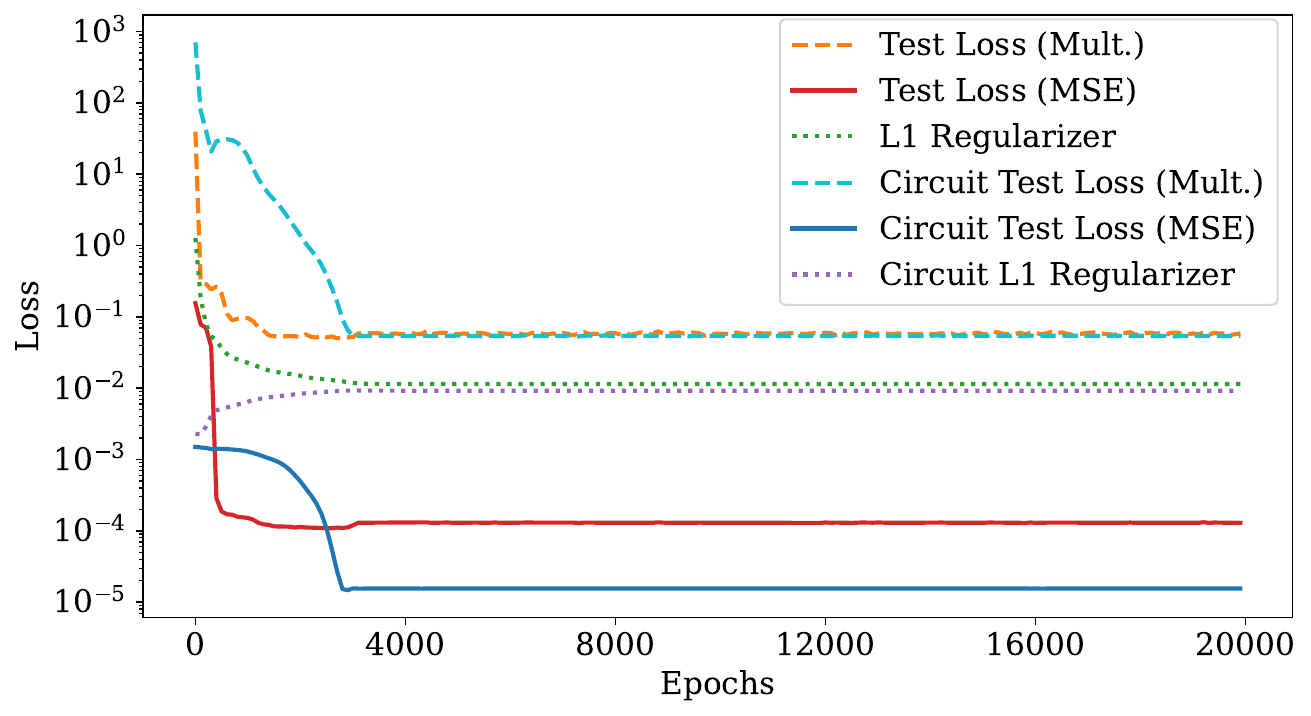}
        \label{subfig:bf-grok-3}
        \caption{Seed 3}
    \end{subfigure}
    \begin{subfigure}{0.49\textwidth}
        \includegraphics[width=\textwidth]{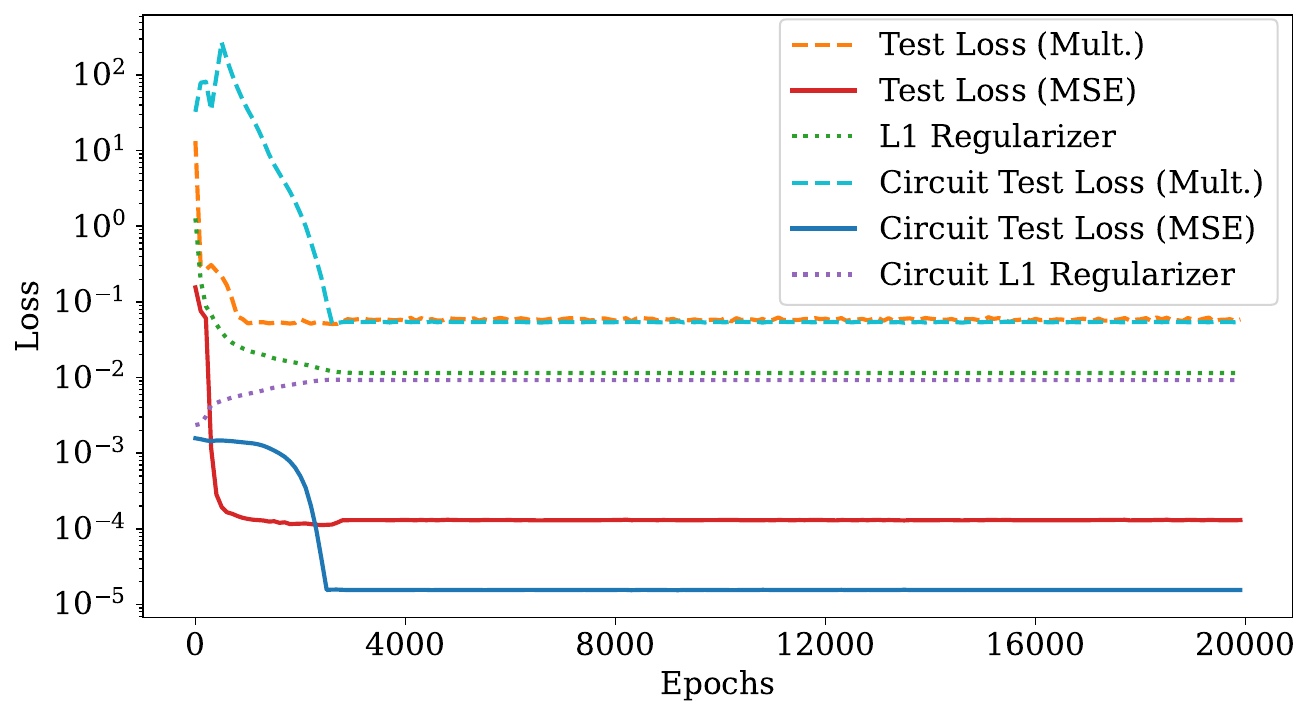}
        \label{subfig:bf-grok-4}
        \caption{Seed 4}
    \end{subfigure}
    \label{fig:bf-circuit-grokking}
    \caption{Bellman-Ford circuit generalization behavior across four additional initializations.}
\end{figure}

\newpage
\subsubsection{Bellman-Ford and Breadth-First Search}
\label{appdx:seeds-parallel}

\begin{figure}[htp]
    \centering
    \begin{subfigure}{0.24\textwidth}
        \includegraphics[width=\textwidth]{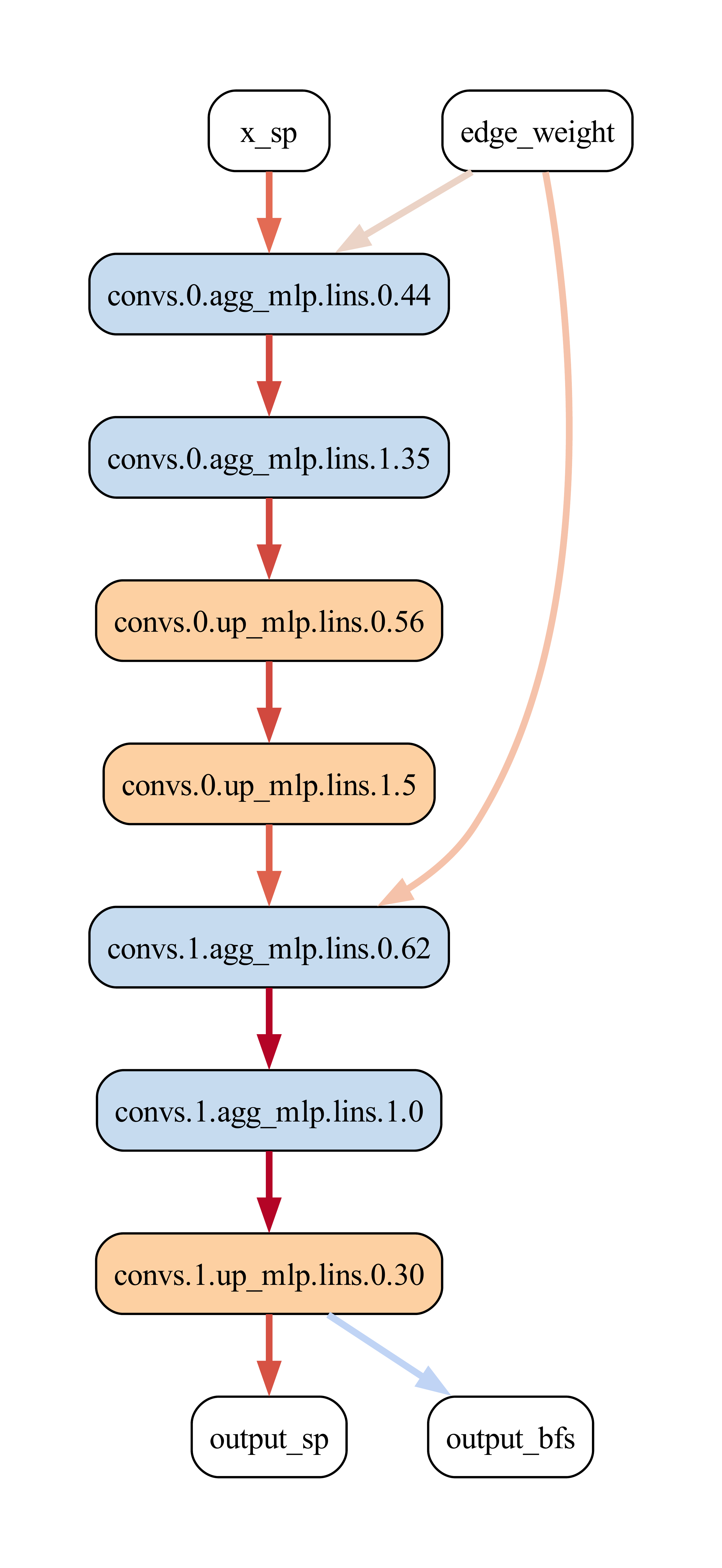}
        \label{subfig:parallel-circuit-1}
        \caption{Seed 1}
    \end{subfigure}
    \begin{subfigure}{0.24\textwidth}
        \includegraphics[width=\textwidth]{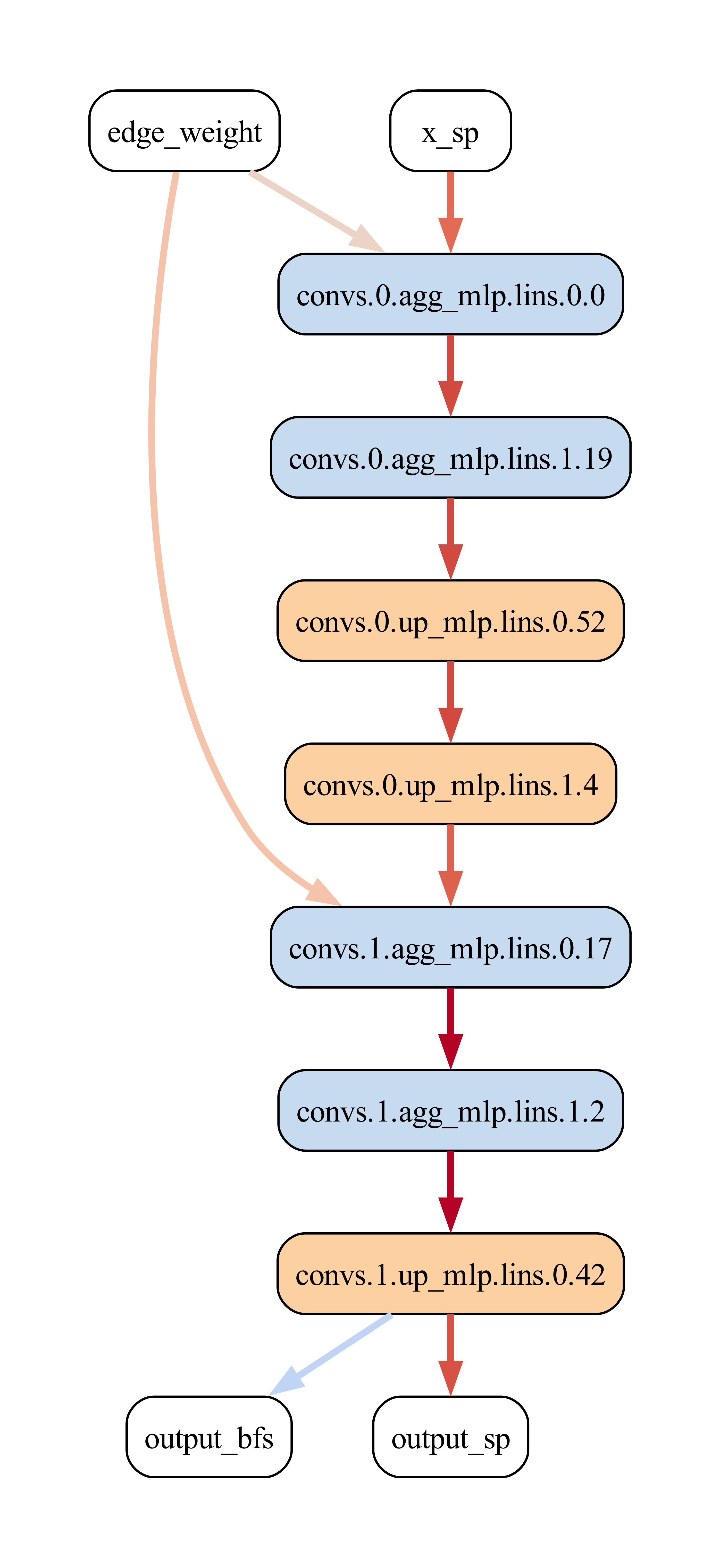}
        \label{subfig:parallel-circuit-2}
        \caption{Seed 2}
    \end{subfigure}
    \begin{subfigure}{0.24\textwidth}
        \includegraphics[width=\textwidth]{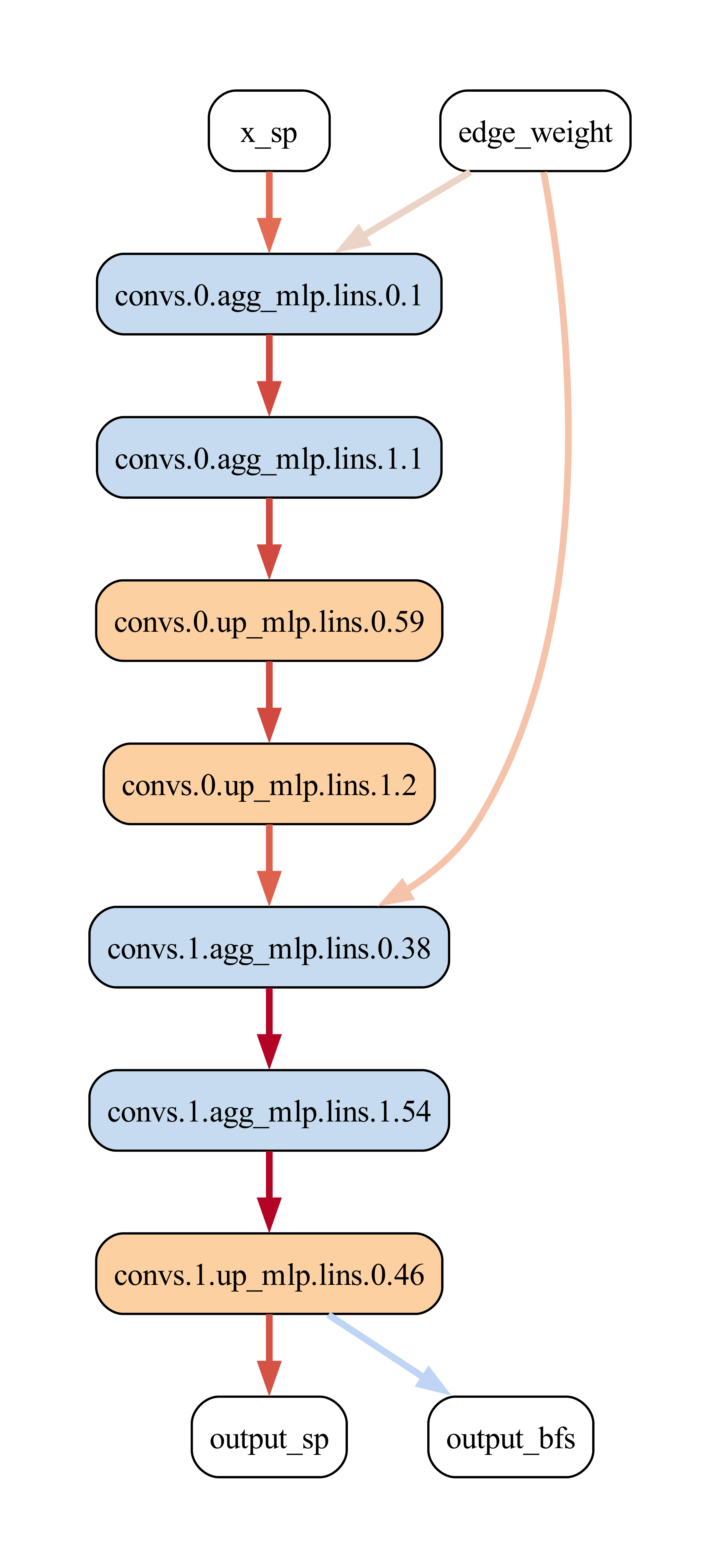}
        \label{subfig:parallel-circuit-3}
        \caption{Seed 3}
    \end{subfigure}
    \begin{subfigure}{0.24\textwidth}
        \includegraphics[width=\textwidth]{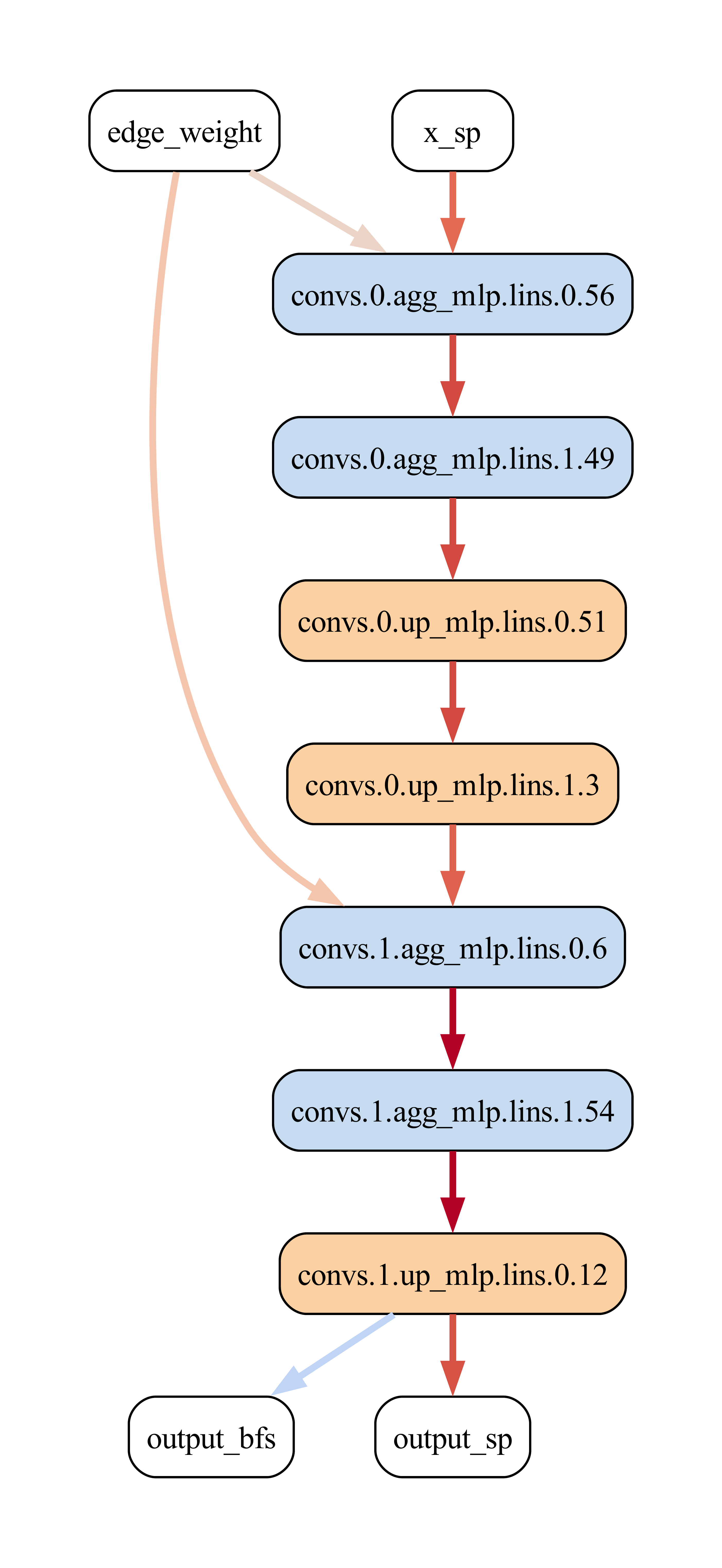}
        \label{subfig:parallel-circuit-4}
        \caption{Seed 4}
    \end{subfigure}
    \label{fig:parallel-circuit-runs}
    \caption{Parallel Bellman-Ford and BFS circuits across four additional initializations.}
\end{figure}

We show circuits extracted across different initializations of the Bellman-Ford and BFS network (\cref{fig:parallel-circuit-runs}). Again, each circuit is functionally the same as the mainline run, but uses different neurons.

\subsection{Parameter Summaries}
\label{appdx:parameter-summaries}

Here we plot parameter summaries for trained Bellman-Ford models from \cref{sec:bellman-ford} and \cref{appdx:no-self-loops}. For readability we only plot results from the mainline run.
\begin{figure}
    \centering
    \includegraphics[width=\linewidth]{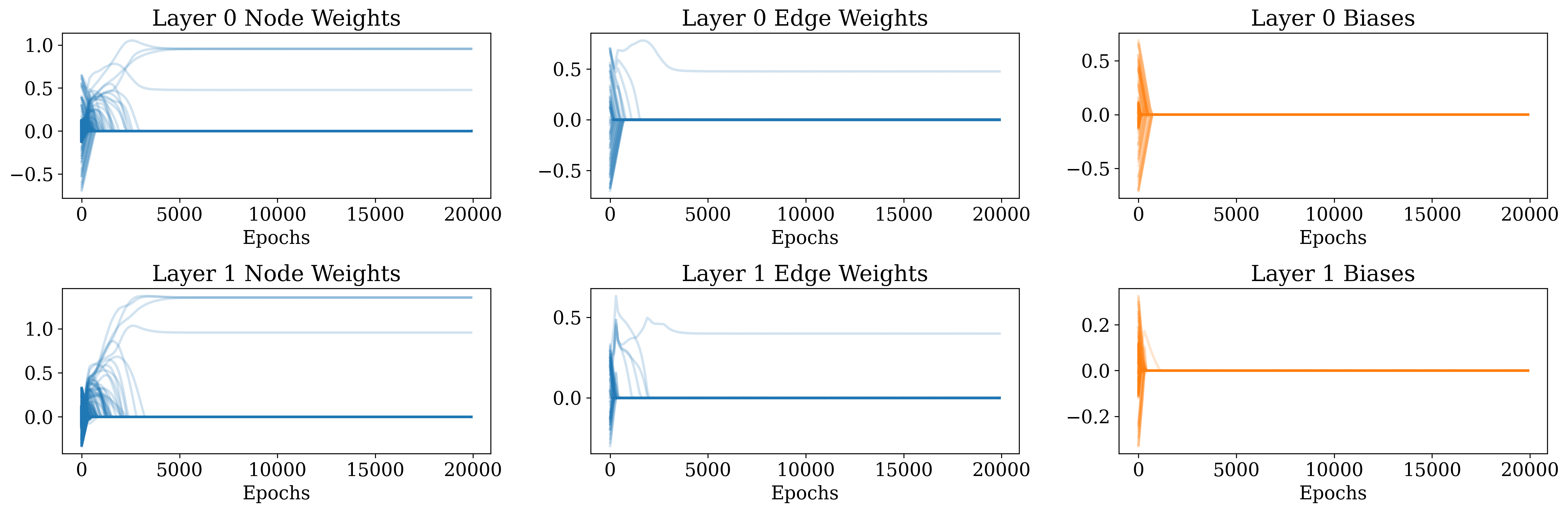}
    \caption{Parameter summary for Bellman-Ford MinAggGNN (\cref{subsec:bellman-ford-circuit}).}
    \label{fig:bellman-ford-params}
\end{figure}
\begin{figure}
    \centering
    \includegraphics[width=\linewidth]{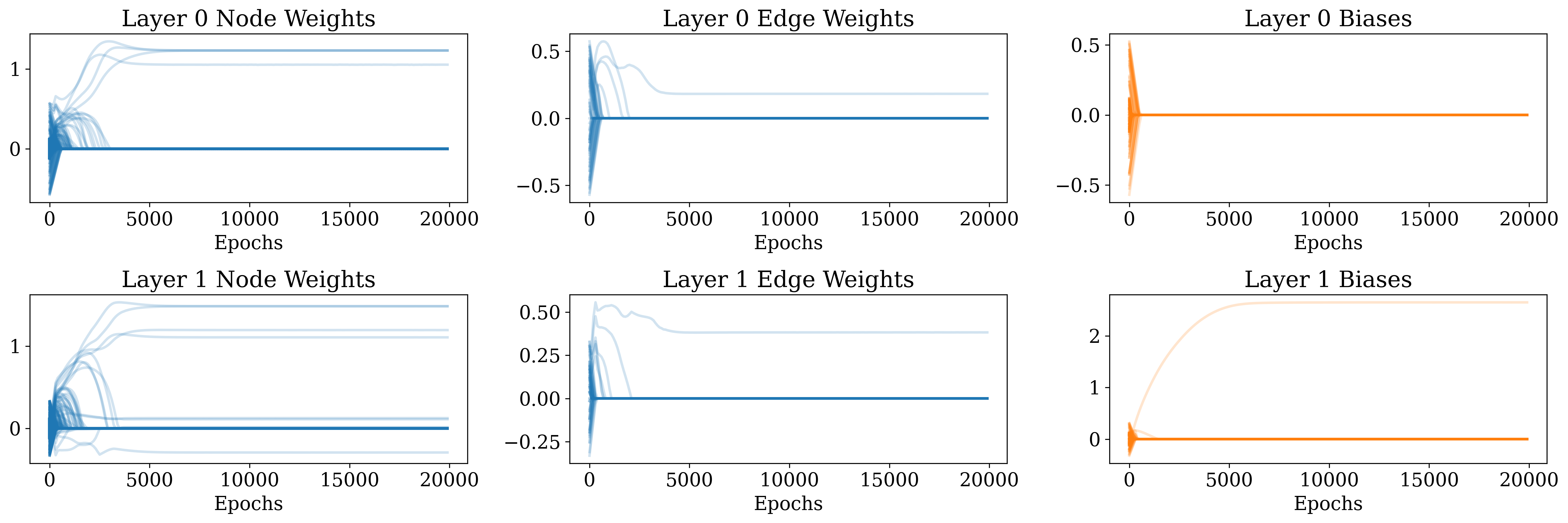}
    \caption{Parameter summary for parallel Bellman-Ford and BFS MinAggGNN (\cref{subsec:bellman-ford-circuit}).}
    \label{fig:parallel-params}
\end{figure}
\begin{figure}
    \centering
    \includegraphics[width=\linewidth]{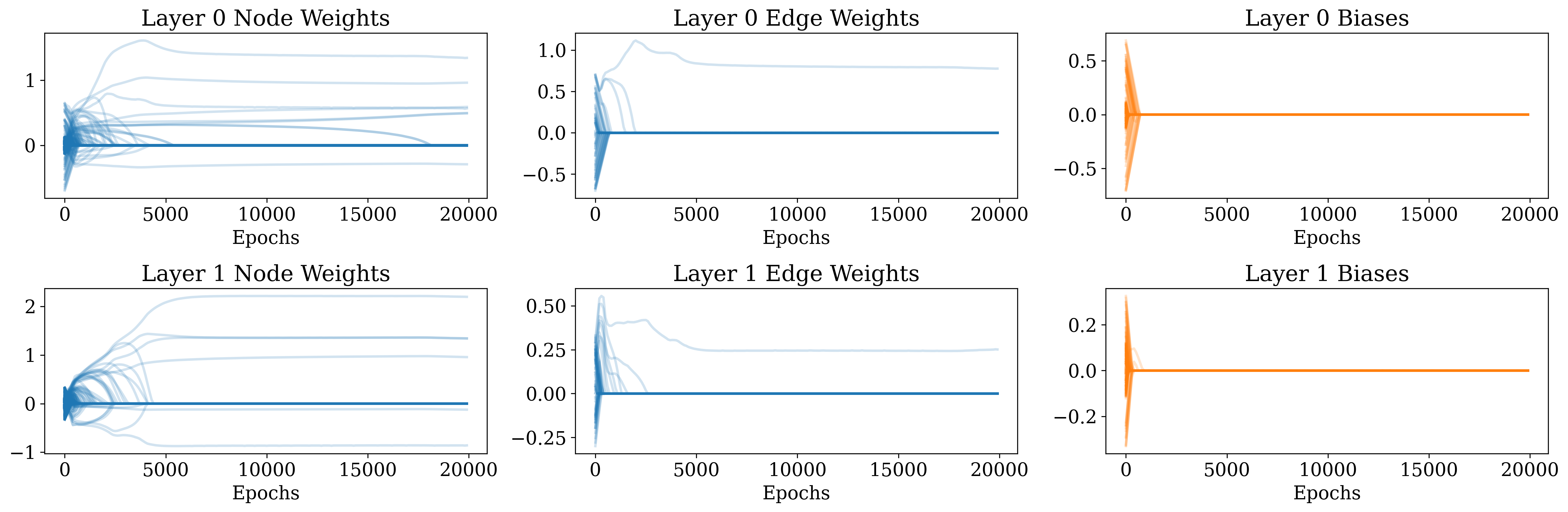}
    \caption{Parameter summary for Bellman Ford MinAggGNN without self-loops (\cref{appdx:no-self-loops}).}
    \label{fig:no-self-loops}
\end{figure}

%% file: sec/appdx/additional-clrs-results.tex
\section{Additional Results on SALSA-CLRS}
\label{appdx:additional-results-clrs}

\begin{figure}[hp]
    \centering
    \includegraphics[width=\linewidth]{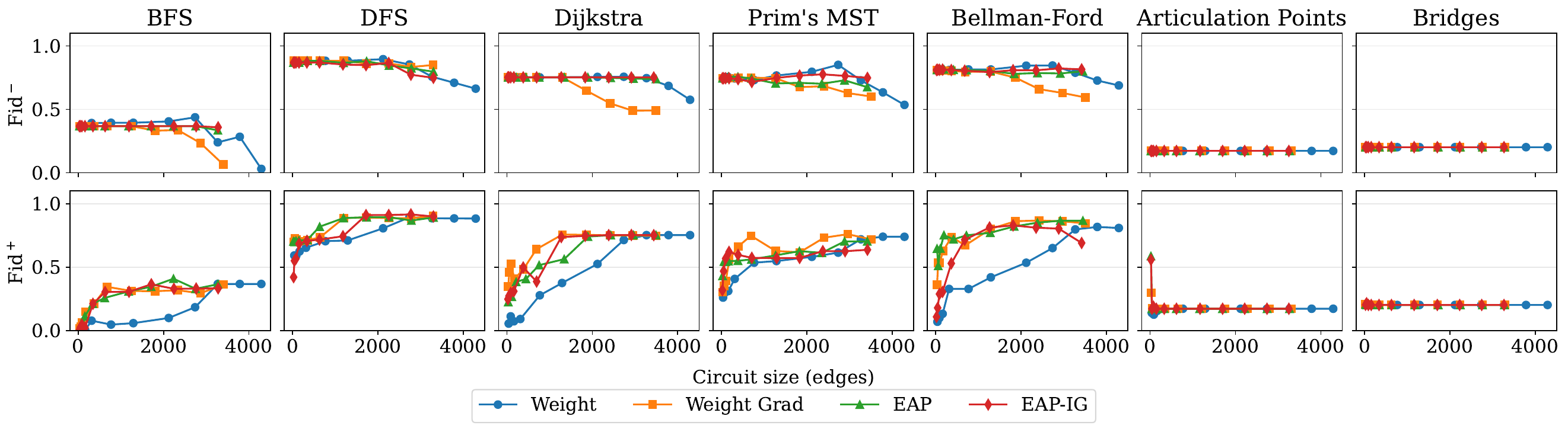}
    \caption{$\text{Fid}^-$ (top) and $\text{Fid}^+$ (bottom) for SALSA-CLRS circuits using Weight, \textsc{WeightGrad}, EAP, and EAP-IG ($m=20$) at $K \in \{10, 25, 50, 100, 250, 500, 1000, 1500, 2000, 2500\}$. For $\text{Fid}^-$, lower indicates a more sufficient circuit. For $\text{Fid}^+$, higher indicates a more necessary circuit.}
    \label{fig:clrs-fidelity}
\end{figure}

\begin{figure}[hp]
    \centering
    \includegraphics[width=\linewidth]{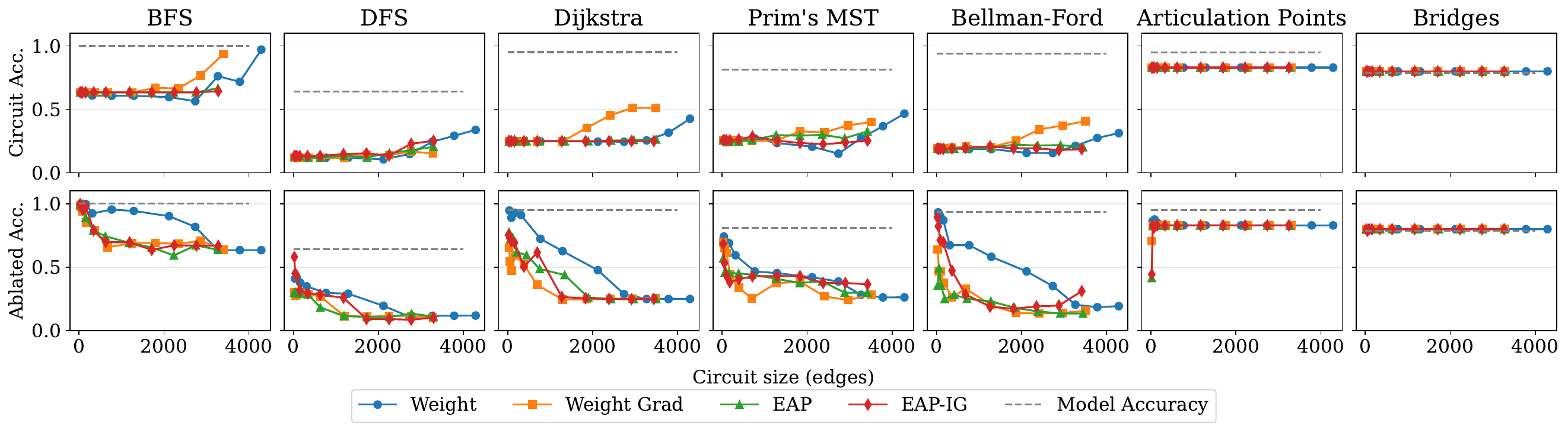}
    \caption{Circuit accuracy (top) and ablated accuracy (bottom) for SALSA-CLRS circuits using Weight, \textsc{WeightGrad}, EAP, and EAP-IG ($m=20$) at $K \in \{10, 25, 50, 100, 250, 500, 1000, 1500, 2000, 2500, 3000\}$. For circuit accuracy, higher is better. For ablated accuracy, lower indicates a necessary circuit.}
    \label{fig:clrs-circuit-accuracy}
\end{figure}

\begin{sidewaysfigure}[hp]
    \centering
    \includegraphics[width=\linewidth]{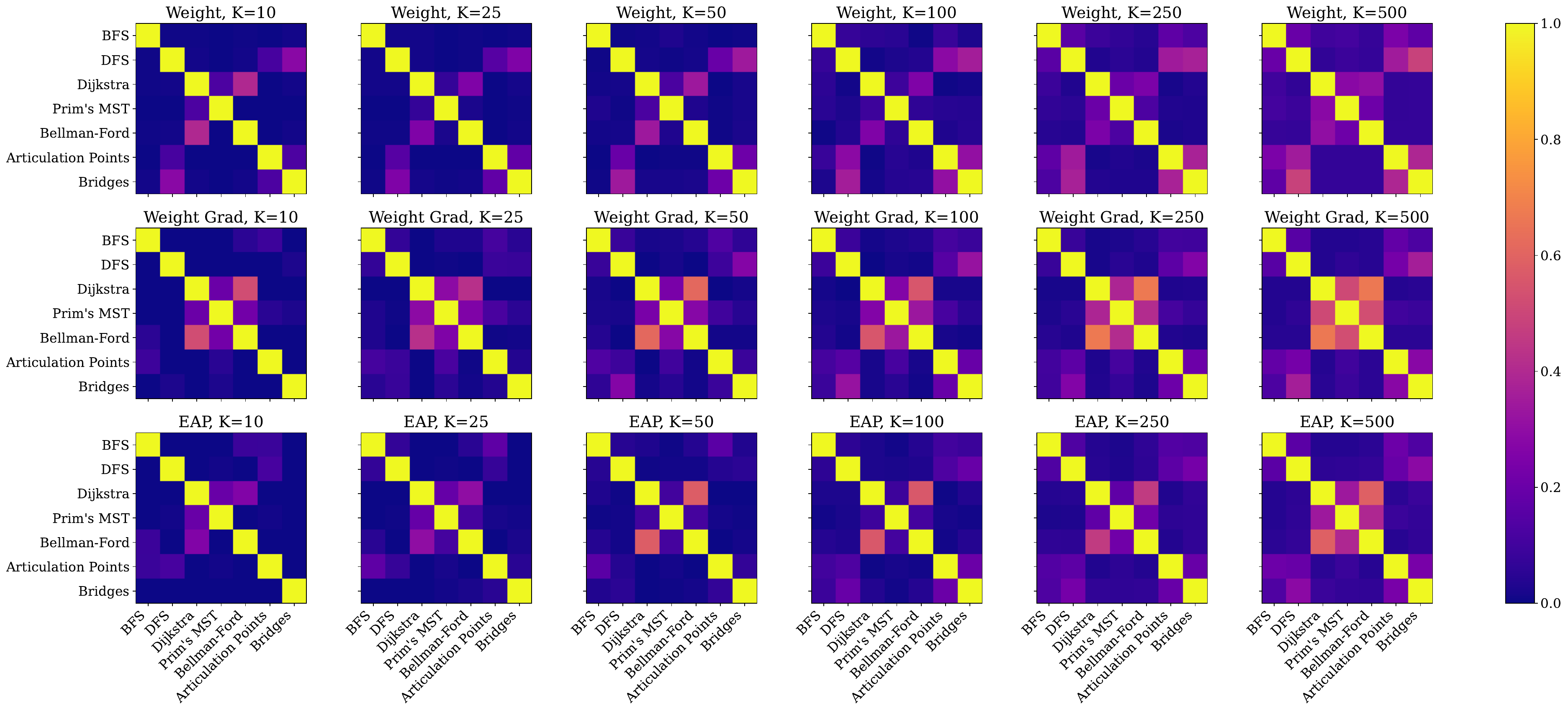}
    \caption{Weighted Jaccard index between CLRS circuits using \textsc{WeightGrad}, EAP, and EAP-IG ($m=20$) at $K \in \{10, 25, 50, 100, 250, 500\}$.}
    \label{fig:clrs-all-overlaps-1}
\end{sidewaysfigure}

\begin{sidewaysfigure}[hp]
    \centering
    \includegraphics[width=\linewidth]{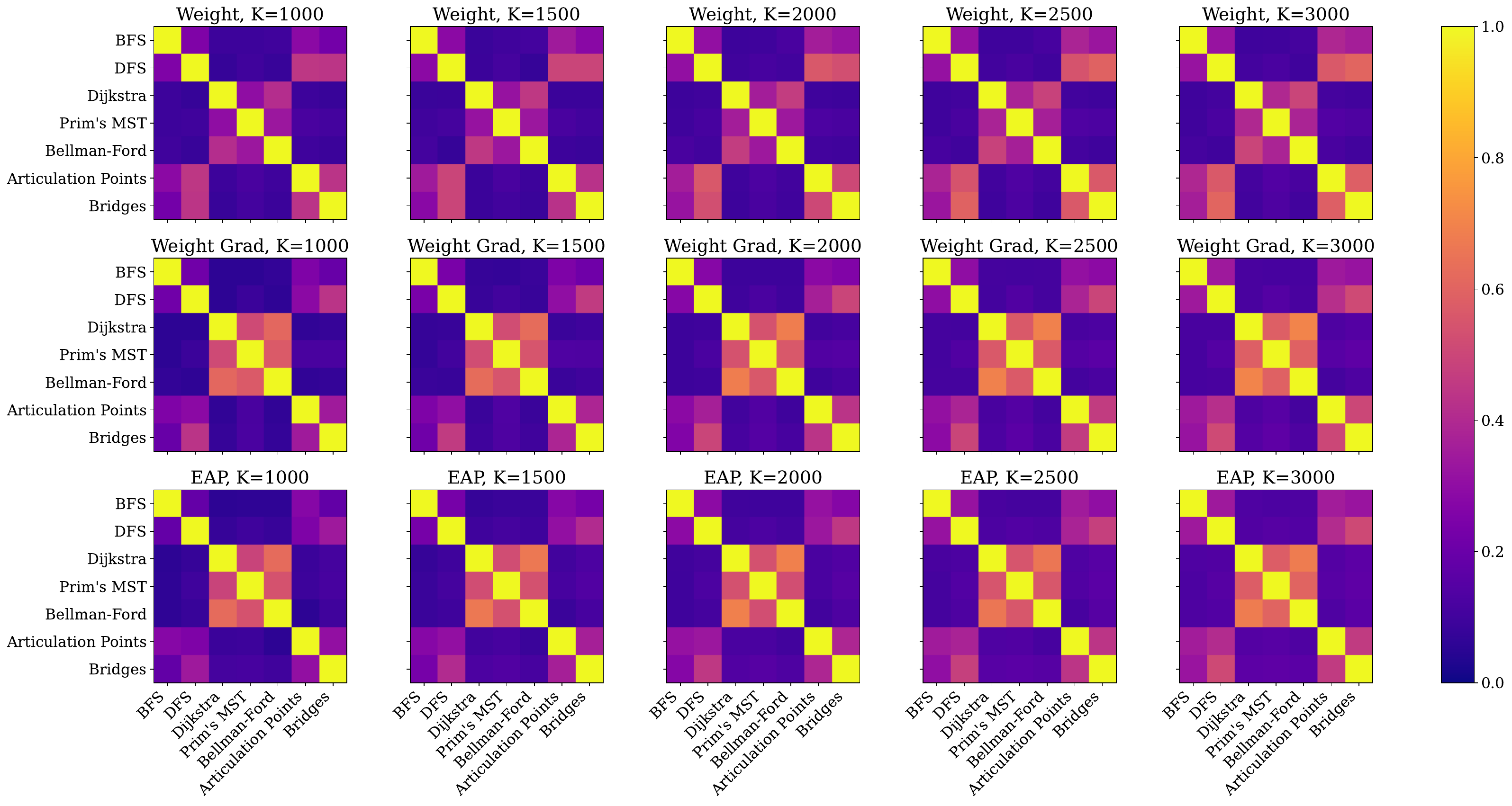}
    \caption{Weighted Jaccard index between CLRS circuits using \textsc{WeightGrad}, EAP, and EAP-IG ($m=20$) at $K \in \{1000, 1500, 2000, 2500, 3000\}$.}
    \label{fig:clrs-all-overlaps-2}
\end{sidewaysfigure}

Here we provide additional results from \cref{sec:salsa-clrs}, including separate fidelity metrics (\cref{fig:clrs-fidelity}), circuit accuracy and ablations on each task (\cref{fig:clrs-circuit-accuracy}), and circuit overlap for each scoring method across all values of $K$ (\cref{fig:clrs-all-overlaps-1}, \cref{fig:clrs-all-overlaps-2}).

%% file: sec/appdx/dataset.tex
\section{Training Data}
\label{appdx:data}

\subsection{Shortest Path Data}
\label{appdx:sp-data}

\begin{figure*}
    \centering
    \includegraphics[width=\linewidth]{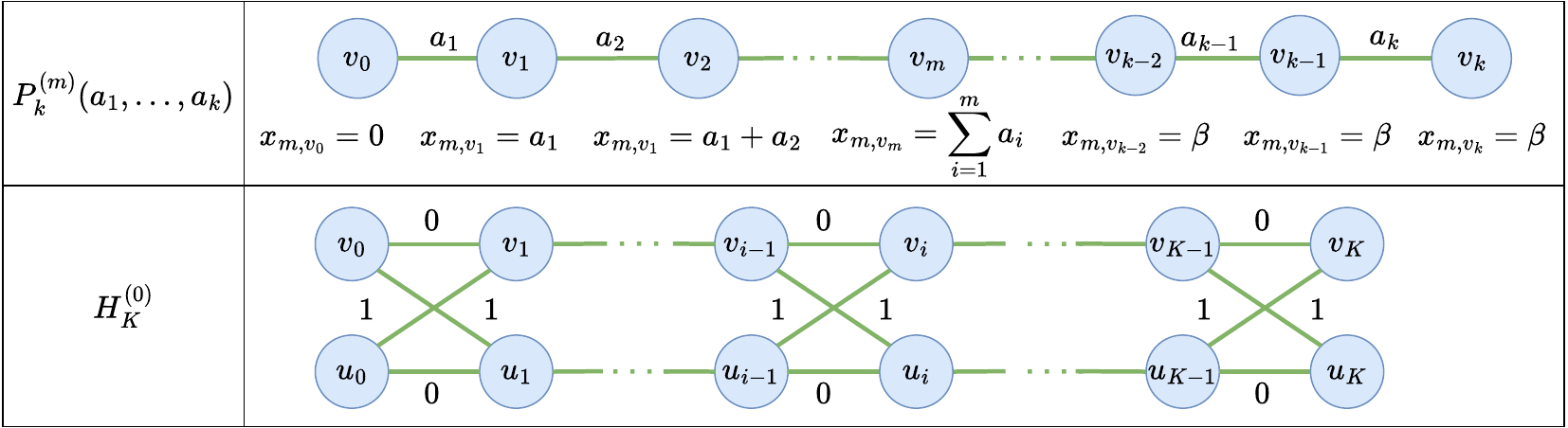}
    \caption{Training graphs for the Bellman-Ford MinAggGNN from~\cite{nerem2025graphneuralnetworksextrapolate}.}
    \label{fig:training}
\end{figure*}

Following~\cite{nerem2025graphneuralnetworksextrapolate}, we construct a training set $\mathcal{G}_{\text{train}}$ from the graphs depicted in \Cref{fig:training} for $K = 2$. It includes all path graphs of the form $P_{K+1}^{(1)}(a,\dots,b,\dots,0)$ for $a, b \in  \{0,1,\dots,2K\}\times\{0,\dots,2K + 1\}$ (where $b$ is the weight of the $K$-th edge). It also includes the graph $H_K^{(0)}$ from \Cref{fig:training} and the special path graphs $P_1^{(0)}(1)$, $P_2^{(1)}(1,0)$. We also include extra path graphs: four three-node path graphs initialized at step zero of Bellman-Ford and four four-node path graphs initialized at step two of Bellman-Ford, each with randomly generated edge weights.

For the out-of-distribution test set, we include a collection of 3 and 4-node cycle graphs; complete graphs on 5 to 200 nodes; and Erdős-Rényi graphs on 5 to 200 nodes with $p=0.5$. To provide extra examples of graphs with unreachable nodes, we also generate binary and ternary trees of depths of 3 and 4. All test graphs have randomly generated edge weights, and the test set contains 300 graphs in total.

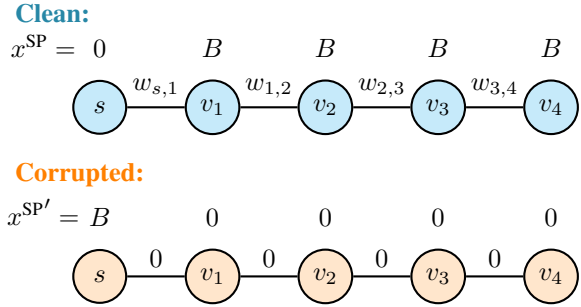
\begin{figure}[ht]
    \centering
    \input{images/bellman-ford/clean_corrupted_data}
    \caption{Example of clean and corrupted data for shortest path length prediction on a path graph.}
    \label{fig:clean_corrupted_data}
\end{figure}

The corrupted dataset includes feature-ablated versions of every graph in the test dataset. An example is given in \cref{fig:clean_corrupted_data}.

\subsection{SALSA-CLRS}
\label{appdx:clrs-data}

We use the datasets described by~\cite{minder2023salsaclrs} for BFS, DFS, Dijkstra's algorithm, and Prim's algorithm. We also generate data for Bellman-Ford, Articulation Points, and Bridges from~\cite{velickovic2022clrs}. (We use implementations of these tasks contained in the codebase for~\cite{minder2023salsaclrs}, but not in the paper.) Each algorithmic task has a training set consisting of 10000 Erdös-Renyi graphs of $n \in \{4,7,11,13,16\}$ nodes, with $p \sim \Unif(\log n / n, 3\log n / n)$ for most tasks except articulation points and bridges, which use $p = \log n / n$. The validation sets consist of 1000 Erdös-Renyi graphs on $n=16$ nodes, and we test out-of-distribution generalization by using test sets of 1000 Erdös-Renyi graphs on $n=80$ nodes with the same ranges for the edge probability $p$. For circuit discovery experiments, we use an additional $128$ unseen graphs drawn from the validation set in batches of 32.

Following~\cite{minder2023salsaclrs}, we use only graph tasks where hints are bound to the given graph structure. For example, in BFS, the hints for each node are pointers to its predecessor in the BFS tree, which must correspond to edges in the actual graph. In contrast, some tasks have nodes maintain pointers to arbitrary nodes. These pointers may not be to immediate neighbors of each node, and hence the pointers may not correspond to actual edges in the graph.

We describe below the exact inputs and outputs for each task, as well as the ablation and loss we use for circuit discovery. In addition to the losses below, we have an additional loss term of $\lambda_{\text{hidden}} = 0.1$ times the MSE between hidden node embeddings on the clean and corrupted data. Note also that every task contains node positions, assigned evenly in [0,1).

\noindent \textbf{Breadth-First Search}~\cite{Moore1959ShortestPath}.
The input consists of a flag for the source node. The output for each node is the edge to its predecessor in the directed BFS tree. The corrupted data removes the source flag and node positions. The loss is given by cross-entropy loss across predicted probabilities for each edge. 

\noindent \textbf{Depth-First Search}~\cite{Moore1959ShortestPath}.
The input consists of a flag for the source node. The output for each node is the edge to its predecessor in the directed DFS tree. The corrupted data removes the source flag and node positions. The loss is given by cross-entropy loss across predicted probabilities for each edge.

\noindent \textbf{Dijkstra's Shortest Path}~\cite{Dijkstra1959}
The input consists of a flag for the source node and edge weights on the graph. The output for each node is the edge to its predecessor in the shortest path from the source. The corrupted data removes the source flag, node positions, and edge weights. The loss is given by cross-entropy loss across predicted probabilities for each edge.

\noindent \textbf{Prim's MST}~\cite{Prim1957ShortestCN}
The input consists of a flag for the source node and edge weights on the graph. The output for each node is the edge to its predecessor in the minimum spanning tree. The corrupted data removes the source flag, node positions, and edge weights. The loss is given by cross-entropy loss across predicted probabilities for each edge.

\noindent \textbf{Bellman-Ford Shortest Path}~\cite{bellman1958routing}
The input consists of a flag for the source node and edge weights on the graph. The output for each node is the edge to its predecessor in the shortest path from the source. The corrupted data removes the source flag, node positions, and edge weights. The loss is given by cross-entropy loss across predicted probabilities for each edge.

\noindent \textbf{Articulation Points}~\cite{cormen2022introduction}
The input consists only of the graph structure and node positions. The task is to classify whether or not each node is an articulation point, i.e. whether its removal would disconnect the graph. The corrupted data removes the node positions. The loss is given by binary cross-entropy loss across predicted probabilities for each node.

\noindent \textbf{Bridges}~\cite{cormen2022introduction}
The input consists only of the graph structure and node positions. The task is to classify whether or not each edge is a bridge, i.e. whether its removal would disconnect the graph. The corrupted data removes the node positions. The loss is given by cross-entropy loss across predicted probabilities for each edge.

\subsection{Designing Corrupted Datasets}
\label{appdx:corr-data}

One major component of edge attribution patching is the design of probing datasets of clean and corrupted inputs. Each corrupted input must be able to align with its corresponding clean input to facilitate computation edge scoring. (This alignment is either between sequences in the LLM case or the graph structure in our case.) At the same time, the corruption must specifically target the behavior of interest. In our case, where the network is explicitly trained to solve specific algorithmic tasks, identifying this behavior of interest is straightforward, but may become more difficult in settings where the algorithmic behavior is part of a more complex downstream task.

In the LLM setting, \citet{dai-etal-2022-knowledge} distill factual information recall into $\langle$Head, Relation, Tail$\rangle$ triples. \citet{wang2023interpretability} isolate indirect object identification by swapping a single name or pronoun in a sentence. We loosely follow this approach of changing tokens without changing structure by ablating features without changing the structure of our corrupted graphs. However, one could consider perturbing the graph structure as well, particularly for CLRS tasks like articulation points or bridges, which are purely structural and do not depend on any features like edge weights or source designations.

%% file: images/bellman-ford/clean_corrupted_data.tex
\begin{tikzpicture}[main node/.style={draw=black,thick,circle,fill=cyan!20!white,minimum size=20pt},
corr node/.style={draw=black,thick,circle,fill=orange!20!white,minimum size=20pt}]
        
        \node[color=cyan!60!black,anchor=west] at (-1.25,1.25) {\textbf{Clean:}};
        \node[main node] (s) at (0,0) {$s$};
        \node[main node] (v1) at (1.5,0) {$v_1$};
        \node[main node] (v2) at (3,0) {$v_2$};
        \node[main node] (v3) at (4.5,0) {$v_3$};
        \node[main node] (v4) at (6,0) {$v_4$};

         \path[draw,thick]
            (s) edge node[above=0pt] (sv1) {$w_{s,1}$} (v1)
            (v1) edge node[above=0pt] (v1v2) {$w_{1,2}$} (v2)
            (v2) edge node[above=0pt] (v2v3) {$w_{2,3}$} (v3)
            (v3) edge node[above=0pt] (v3v4) {$w_{3,4}$} (v4);

        \node[above=5pt of s] (clean) {$0$};
        \node[above=5pt of v1] {$B$};
        \node[above=5pt of v2] {$B$};
        \node[above=5pt of v3] {$B$};
        \node[above=5pt of v4] {$B$};
        \node at (-.75,.75cm+2pt) {$x^{\text{SP}}=$};

        \node[color=orange,anchor=west] at (-1.25,-.9) {\textbf{Corrupted:}};
        \node[corr node] (s') at (0,-2.25) {$s$};
        \node[corr node] (v1') at (1.5,-2.25) {$v_1$};
        \node[corr node] (v2') at (3,-2.25) {$v_2$};
        \node[corr node] (v3') at (4.5,-2.25) {$v_3$};
        \node[corr node] (v4') at (6,-2.25) {$v_4$};

         \path[draw,thick]
            (s') edge node[above=0pt] (sv1') {$0$} (v1')
            (v1') edge node[above=0pt] (v1v2') {$0$} (v2')
            (v2') edge node[above=0pt] (v2v3') {$0$} (v3')
            (v3') edge node[above=0pt] (v3v4') {$0$} (v4');

        \node[above=5pt of s'] (corr) {$B$};
        \node[above=5pt of v1'] {$0$};
        \node[above=5pt of v2'] {$0$};
        \node[above=5pt of v3'] {$0$};
        \node[above=5pt of v4'] {$0$};
        \node at (-.75,-1.5cm+3pt) {${x^{\text{SP}}}'=$};

    \end{tikzpicture}

%% file: sec/appdx/training.tex
\section{Model Training and Hardware}
\label{appdx:training}

Computations for \cref{sec:bellman-ford} and \cref{appdx:no-self-loops} were performed on an NVIDIA RTX A6000 Laptop GPU using PyTorch~\cite{paszke2019pytorch} and PyTorch-Geometric~\cite{Fey2019FastGR}. For each MinAggGNN, $f^{(\ell)}_{\Agg}$ and $f^{(\ell)}_{\Up}$ are implemented as two-layer MLPs with a hidden dimension of 64. The intermediate dimension of each network is 8. We train each model using AdamW~\cite{loshchilov2018decoupled} with a learning rate of $\gamma = 0.001$ for 20000 epochs using full batches, a default weight decay of $\lambda = 0.01$, and an $L_1$ regularization term of $\eta = 0.001$.

Because the parallel Bellman-Ford and BFS experiment uses the same training and testing sets as the Bellman-Ford experiment, the positive and negative reachability classes are imbalanced. (About 91.24\% of training nodes and 81.98\% of test nodes are reachable.) Therefore, we use class weighting so that the positive and negative classes have equal weight during training. We additionally scale the BCE loss by a factor of 25 during training, as we observe that the MSE and $L_1$ terms dominate training otherwise.

Circuit analysis for \cref{sec:salsa-clrs} were performed on the same NVIDIA RTX A6000 Laptop GPU. The model was trained on an HPC cluster using 7 NVIDIA A100 GPUs, each with 80GB of memory. Each task is trained on a single GPU and gradients are synced using PyTorch's distributed data parallel module. The model was trained for 100 epochs using a minibatch size of 32 using AdamW with an initial learning rate of $\gamma_0 = 0.001$, with cosine annealing~\cite{loshchilov2017sgdr} down to $\gamma_{100} = 1 \times 10^{-5}$. To encourage sparsity, we began with an initial weight decay of $\lambda = 0.1$, which we decreased to 0 by cosine annealing, and an initial $L_1$ regularization term of $\eta_0 = 0$ which we increased to $\eta_{100} = 0.001$ by $\eta_t = \eta_{\max}\left(1-\cos \frac{\pi t}{T}\right)$.